\documentclass[manuscript,screen]{acmart}

\usepackage{amsmath}
\usepackage{amsthm}
\usepackage{centernot}
\usepackage{longtable}
\usepackage{placeins}
\usepackage{enumitem} 
\setcopyright{none}

\newtheorem{proposition}{Proposition}
\newcommand{\R}{\mathbb{R}}
\newcommand{\E}{\mathbb{E}}
\newcommand{\Prob}{\mathbb{P}}
\newcommand{\Var}{\mathrm{Var}}
\newcommand{\ind}{\mathbf{1}}
\DeclareMathOperator{\expit}{expit}

\DeclareMathOperator{\Bernoulli}{Bernoulli}
\DeclareMathOperator{\Uniform}{Uniform}
\newcommand\independent{\protect\mathpalette{\protect\independenT}{\perp}}
\def\independenT#1#2{\mathrel{\rlap{$#1#2$}\mkern2mu{#1#2}}}

\begin{document}

\title{Targeted Learning for Inference on Data Fairness}

\author{Joseph Rudoler}
\affiliation{%
  \institution{University of Pennsylvania}
  \country{USA}
}
\email{jrudoler@wharton.upenn.edu}

\author{Alexander Asemota}
\affiliation{%
  \institution{University of California, Berkeley}
  \country{USA}
}
\email{asemota.alexander@gmail.com}

\author{Giles Hooker}
\affiliation{%
  \institution{University of Pennsylvania}
  \country{USA}
}
\email{ghooker@wharton.upenn.edu}

\renewcommand{\shortauthors}{Rudoler et al.}

\begin{abstract}
  As algorithmic decision making has become increasingly ubiquitous even in high-stakes settings, significant effort has been invested in defining, detecting, and eliminating unfair model-driven outcomes.
  Prior work has largely focused on estimating the fairness properties of a given predictive model, and quantifying uncertainty based on variability in a fixed model's performance across evaluation datasets.
  Alternatively, we can also ask about uncertainty that the model inherits from its training data; here we evaluate \emph{data fairness} -- fairness in the data generating process used for training.
  We perform inference on data fairness using targeted learning, a flexible framework for nonparametric inference, and derive estimators for demographic parity, equal opportunity, and conditional mutual information.
  Simulations show that our data fairness estimators improve confidence-interval coverage and reduce false detections of population unfairness relative to fixed-model inference.
  We also demonstrate their application to real data.
\end{abstract}



\maketitle

\section{Introduction}
Modern machine learning has considerably amplified the scale and scope of data-driven decision-making even in high-stakes settings. This has led to a literature on evaluating the fairness of these powerful models and comparing their predictions across different demographic groups \cite{catonFairnessMachineLearning2024}.
To date, however, most fairness research has focused on \emph{model fairness}, which evaluates the fairness of a single fitted model or decision rule.
This is a natural perspective, since in practice a particular model is deployed and its predictions are the basis for decisions.
Uncertainty estimates for model fairness ask about how much sampling variability in the data used to evaluate fairness affects our conclusions about a specific model; they ignore additional uncertainty that the model inherits from sampling variability in the training data.
Fairness concerns in machine learning models have frequently been attributed to biases inherited from their training data \cite{barocasBigDatasDisparate2014, lumPredictServe2016}.
Prior work has shown that even fairness-adjusted models can perpetuate bias if the outcome is biased \cite{kallusResidualUnfairnessFair2018}.
Consequently, current single-model approaches fail to address whether apparent unfairness can be traced to fairness violations in the data generating mechanism already present during data collection; they are also agnostic to whether unfairness is an artifact of fitting on a small data sample \cite{chenWhyMyClassifier2018}.
Here we provide inferential methods for \emph{data fairness} -- fairness in the data-generating process used to learn decision rules \cite{tangWhatisHowtoFairness2023}.
Beyond providing tools to address these questions, we also believe it is important to provide terminology to differentiate these sources of uncertainty, to be able to clearly differentiate which is relevant in any given circumstance.

Distinguishing and accounting for the correct sources of uncertainty is important when providing real-world interpretations of models. \citet{huangFairnessTestingAlgorithmic2026} study model-fairness audits of fixed insurance-pricing algorithms and show that inappropriate variance calculations assuming sampling variability can distort inference, inflating uncertainty and obscuring discrimination in proxy-discrimination tests. On the other hand, if we are interested in whether we can expect a particular data source to consistently result in models that violate fairness metrics -- for example if we expect frequent model updates -- we are explicitly interested in data fairness and may seek to add or remove features to reduce the bias in an idealized model.
This can also arise rather naturally even without an explicit model; consider a university that wants to know whether its admissions process is unfair to a particular group.
The university uses application materials $X$ to make admissions decisions predicting an outcome $Y$, and asks whether decisions based on $X$ are unfair with respect to group membership $G$.
Because no fixed algorithm is under scrutiny, this is not a question of model fairness.
It concerns only the map from the chosen set of features $X$ to the outcome $Y$, without concern for the specific mechanisms at play, in contrast to causal approaches to fairness \cite{khademiFairnessAlgorithmicDecision2019}.
We formalize data fairness as a fairness criterion that is a functional of the observed data-generating distribution alone.
Such a criterion asks whether decisions under the true relationship between $X$ and $Y$ nonetheless differ systematically across groups.
Because it pertains to the data used to inform decisions, a violation can affect models or decision rules built from that data.

Data fairness can be expressed as a problem of classical statistical inference: we would like a confidence interval for a fairness violation in the underlying population from which we have a sample. Our challenge is to provide these quantities when the conditional distribution of $Y$ is described nonparametrically, and estimated using methods from machine learning.
Here we rely on \emph{targeted learning} (TL) to estimate the uncertainty associated with that estimate.
Targeted learning is a framework for nonparametric inference which couples machine learning predictions with a bias-correction step based on efficient influence functions (EIFs), yielding asymptotically unbiased estimators which admit valid confidence intervals \cite{vanderlaanTargetedLearningCausal2011,hinesDemystifyingStatisticalLearning2022}.
Since the estimand is defined directly on the data-generating distribution, this lets us target fairness properties of the data itself, rather than of a single fitted model, while still quantifying uncertainty.

Our contributions are as follows:
\begin{itemize}[noitemsep,topsep=0pt]
  \item We formalize \emph{data fairness}, mathematically, as a function of the data generating process rather than a particular model or decision rule.
  \item We use targeted learning to derive estimators and valid confidence intervals for several data fairness metrics, including demographic parity, equal opportunity, and conditional mutual information (CMI).
  \item We demonstrate when our estimators improve coverage over naive alternatives and identify limitations with small samples and misspecified nuisance models.
  \item We demonstrate how the choice of a set of data features affects estimates of data fairness.
\end{itemize}

The remainder of this paper is structured as follows: Section \ref{sec:inference} introduces data fairness and the application of targeted learning to deriving appropriate estimators.
In Section \ref{sec:sims}, we perform simulations to evaluate the properties of our derived estimators, and in Section \ref{sec:data}, we apply our estimators to two real-world datasets.
Finally, in Section \ref{sec:discussion}, we discuss the implications of this work along with potential future work.

\paragraph{Related Work}
Statistical inference for fairness auditing has been developed for fixed prediction models.
\citet{cherianStatisticalInferenceFairness2024} construct simultaneous bootstrap bounds on performance disparities across collections of subpopulations, allowing auditors to flag underperformance and certify acceptable performance.
\citet{gaoReliableFairnessAuditing2026} develop semi-supervised inference for fairness metrics of a fixed classifier, combining labeled and unlabeled audit data through calibrated outcome imputation.
\citet{tramerFairTestDiscoveringUnwarranted2016} propose a method that discovers and tests unwarranted associations between application outputs and protected attributes, accounting for multiple comparisons and allowing explanatory factors.
These procedures quantify uncertainty about a fitted rule, conditional on its training data; our estimands instead concern the population regression $D(x)=\Prob(Y=1\mid X=x)$, and our influence-function corrections account for estimation of that regression.

Other work moves beyond a single fitted model and examines fairness across random samples or stochastic optimization runs.
\citet{royFairnessUncertaintyQuantification2023} derive asymptotic distributions and an online bootstrap procedure to quantify fairness uncertainty for linear classifiers trained with fairness-aware stochastic optimization.
\citet{ganeshImpactMachineLearning2023} empirically investigate variation in group fairness across neural-network training runs, separating the effects of initialization and training-data order.
\citet{chenWhyMyClassifier2018} is close in spirit to our work, examining how sampling variability in finite training data and unmeasured predictive variables may contribute to unfairness.
They define expected disparities in group-specific prediction costs by averaging over training samples drawn from the population, and decompose these disparities into bias, variance, and noise to assess whether additional observations or predictive features could reduce them.
Whereas we study the distributions of the population regression directly, their analysis depends on the learning algorithm, model class, and training-sample size.
They provide CLT-based tests and confidence intervals which evaluate disparities for a fixed model; they recommend resampling methods such as bootstrapping to address the uncertainty induced by training data.
This approach requires resampling the training data and refitting the model in each bootstrap replicate, which may be computationally expensive.

In a thoughtful review of fairness notions and their respective theories of justice, \citet{tangWhatisHowtoFairness2023} explicitly distinguish ``fairness w.r.t. data generating process'' from ``fairness w.r.t. predicted outcome''.
The former is motivated by prior work discussing how data can encode both \emph{societal} bias as well as \emph{statistical} bias \cite{mitchellAlgorithmicFairnessChoices2021}.
They formulate data fairness criteria using observational causal inference, requiring corresponding causal identifiability assumptions.
Here we take a more flexible approach, avoiding causal assumptions and instead relying on nonparametric statistical inference to quantify uncertainty about fairness in the data-generating process.

We estimate fairness properties of the observed population used to train a model.
Related work mitigates discrimination by modifying training data before model fitting, including label modification and reweighting \cite{kamiranDataPreprocessingTechniques2012} and optimized data transformations \cite{calmonOptimizedDataPreProcessing2017}.

This work also does not consider the cost of enforcing fairness; in practice, there is likely a tradeoff between fairness and other objectives like accuracy and its downstream benefits \cite{corbett-daviesAlgorithmicDecisionMaking2017, xianFairOptimalClassification2023, liangAlgorithmDesignFairnessAccuracy2024,liuInferenceAlgorithmicFairnessAccuracy2025}.
\citet{liuInferenceAlgorithmicFairnessAccuracy2025} use related debiased machine learning methods to estimate the attainable tradeoffs between group-specific prediction losses and disparities in those losses across decision rules using a given feature set.
Their target describes opportunities to improve fairness and accuracy by choosing a different rule; ours measures parity and opportunity disparities of the population regression $D(x)=\Prob(Y=1\mid X=x)$.

\section{Inference for Data Fairness}
\label{sec:inference}

A model-based approach to fairness ties inference to one specific fit, ignoring uncertainty due to both sampling variability in the training data and stochasticity in model-fitting.
Defining estimands on the data-generating distribution instead lets us draw conclusions about the underlying process, and hence about any accurate model or decision rule built from the data it generates.
Of course, the data-generating distribution is unknown, so we must estimate it and account for the uncertainty that entails.
Targeted learning \cite{vanderlaanTargetedLearningCausal2011,hinesDemystifyingStatisticalLearning2022} is a flexible framework for nonparametric inference that provides a recipe for doing so.

\subsection{Setup}
\label{sec:setup}

We observe $n$ independent and identically distributed copies of $O = (X, G, Y) \sim P,$ where $X$ collects the features available to a decision maker, $G$ is a sensitive attribute (typically group membership), and $Y$ is an outcome.
We take both $G$ and $Y$ to be binary throughout, $G, Y \in \{0,1\}$, as is standard for the fairness metrics we build on, although extensions to more complex situations are feasible.
Additional conditions for inference are given in Appendix \ref{app:assume}.
As $P$ is generally unknown, we attempt to estimate it flexibly using a nonparametric statistical model $\mathcal{M}$. We consider estimands which can be expressed as functionals $\Psi: \mathcal{M} \to \R$.

\paragraph{Probability distributions.}
Every estimand we study is a functional of the joint conditional law,
\begin{equation}
  p(g,y \mid x) = \Prob(G=g, Y=y \mid X=x),
  \label{eq:joint}
\end{equation}
a distribution over the four states of the binary pair. This distribution describes the relationship between the sensitive attribute and the outcome, conditional on the features.
The quantities that appear in the individual influence functions in this study are all margins or slices of \eqref{eq:joint}:
\begin{align}
  D(x)      & = \Prob(Y=1 \mid X=x)       = \textstyle\sum_{g} p(g,1 \mid x)
  && \text{(Outcome probability)}, \\
  \pi(x)    & = \Prob(G=1 \mid X=x)       = \textstyle\sum_{y} p(1,y \mid x)
  && \text{(Group-membership probability)}, \\
  \rho_g(x) & = \Prob(Y=1, G=g \mid X=x)  = p(g,1 \mid x)
  && \text{(Joint outcome--group probability)}.
\end{align}

\paragraph{Data fairness.}
We call a fairness criterion a \emph{data fairness} criterion when it is a functional of the data distributions alone.
It can be stated and estimated without reference to any fitted model or predictions.
Because such a criterion describes the data-generating process, a violation propagates to any decision rule learned from that data.
An example criterion, which we carry through the rest of this section, is \emph{probabilistic demographic parity}, which requires that the average conditional probability of a positive outcome is the same in both groups.
This holds when the estimand
\begin{equation}
  \Psi_{\mathrm{pDP}}(P) = \E[\,D(X) \mid G=1\,] - \E[\,D(X) \mid G=0\,],
  \label{eq:demographic_parity}
\end{equation}
is zero.
The specifier \emph{probabilistic} distinguishes it from the thresholded form we will study in Section \ref{sec:thresholded}, which compares rates of binary decisions rather than average probabilities.

\paragraph{From estimand to inference.}
Now we have an estimand defined on $P$. In practice, though, we rarely know $P$ or its margins, so we instead estimate it using flexible learning methods.
Consider a flexible estimator $\hat{P}$ of the data-generating distribution $P$ (or one of its components), which we will also refer to as the \emph{nuisance} parameters.
The plug-in $\Psi(\hat{P})$ -- the fairness estimand evaluated at the fitted distribution -- accrues bias from the estimation of $P$.
The key insight of targeted learning is that the influence function $\phi$, essentially the derivative of $\Psi$ with respect to small changes in $P$, captures the first-order part of that bias \cite{hinesDemystifyingStatisticalLearning2022,kennedySemiparametricDoublyRobust2023}.
This should be intuitive: the bias depends on how sensitive the estimand $\Psi$ is to misspecification of the nuisance $\hat{P}$.
An important property of influence functions is that they are mean-zero under $P$, that is $P[\phi(O;P)] = 0$. This normalization reflects the fact that perturbing $P$ must preserve the total probability mass.

Applying a von Mises expansion \cite{misesAsymptoticDistributionDifferentiable1947} of $\Psi$ to first order about $P$ gives
\begin{equation}
  \Psi(\hat{P}) - \Psi(P) = -P[\phi(O;\hat{P})] + R(\hat{P},P),
  \label{eq:remainder}
\end{equation}
which defines the remainder $R(\hat{P},P)$ as everything the first-order term does not capture.
Adding and subtracting empirical averages splits that first-order term into three pieces that can be controlled separately,
\begin{equation}
  \Psi(\hat{P}) - \Psi(P) = \underbrace{P_n[\phi(O;P)]}_{\text{leading term}} - \underbrace{P_n[\phi(O;\hat{P})]}_{\text{plug-in bias}} + \underbrace{(P_n - P)\big[\phi(O;\hat{P}) - \phi(O;P)\big]}_{\text{empirical process}} + \underbrace{R(\hat{P},P)}_{\text{remainder}},
  \label{eq:decomposition}
\end{equation}
where $P_n$ is the empirical distribution of the evaluation sample, and we drop $P[\phi(O;P)] = 0$.

The leading term is an average of i.i.d. mean-zero variables, so it obeys the central limit theorem and supplies the limiting distribution.
We can modify the estimator by adding the empirical mean of $\phi(O;\hat{P})$ over the evaluation sample: replacing $\Psi(\hat{P})$ with $\bar{\Psi}(\hat{P}) = \Psi(\hat{P}) + P_n[\phi(O;\hat{P})]$ cancels the second term of \eqref{eq:decomposition}.
Sample splitting keeps $\hat{P}$ independent of the evaluation data, so the empirical process term behaves like a product of sampling fluctuations of order $n^{-1/2}$ and the $\mathcal{L}_2$ error in the estimated influence function.
When the influence function is consistently estimated, that second factor converges to zero, making the product $o_P(n^{-1/2})$.
The remainder requires the further assumptions detailed in Appendix \ref{app:assume}.
An alternative is targeted maximum likelihood estimation (TMLE), which updates the fitted distribution so that its empirical influence-function mean is approximately zero, then evaluates the estimand at that targeted distribution \cite{vanderlaanTargetedLearningCausal2011}. We use this substitution approach for CMI to preserve its nonnegativity; the construction and its conditions are given in Appendix \ref{app:cmi_eif}.
For the additive correction with independently fitted nuisances, these conditions give the following result:

\begin{proposition}[asymptotic linearity]
  \label{prop:linear}
  Suppose the nuisances are fit on an independent split and the resulting influence function is $\mathcal{L}_2$-consistent, $\|\phi(\,\cdot\,;\hat{P}) - \phi(\,\cdot\,;P)\| = o_P(1)$; the remainder of \eqref{eq:decomposition} satisfies $R(\hat{P},P) = o_P(n^{-1/2})$; and the positivity conditions of Appendix \ref{app:assume} hold.
  Then $\bar{\Psi}(\hat{P})$ is asymptotically linear with influence function $\phi(\,\cdot\,;P)$,
  $$\sqrt{n}\left(\bar{\Psi}(\hat{P}) - \Psi(P)\right) \xrightarrow{d} N\!\left(0,\, \Var_P[\phi(O;P)]\right),$$
  If $0 < \Var_P[\phi(O;P)] < \infty$, then $\bar{\Psi}(\hat{P}) \pm z_{1-\alpha/2}\,\hat{\sigma}/\sqrt{n}$, with $\hat{\sigma}^2$ the sample variance of $\phi(O;\hat{P})$, is an asymptotically valid $1-\alpha$ confidence interval.
\end{proposition}

This result permits us to construct confidence intervals for any data fairness estimand $\Psi$ for which we can derive an influence function and satisfy the conditions of Proposition \ref{prop:linear}.

\subsection{Demographic Parity}
\label{sec:parity}

Fairness metrics are traditionally defined relative to a fitted model and its predictions.
Demographic parity, for instance, requires \cite{hardtEqualityOpportunitySupervised2016}
$$\Prob(\hat{y} = 1 \mid G=1) = \Prob(\hat{y} = 1 \mid G=0),$$
where $\hat{y}$ is a model's prediction; this is a property of the specific model that produced $\hat{y}$.
To characterize the data instead, we replace the model output $\hat{y}$ with a functional of the conditional law of $Y \mid X$, so that inferences about parity become inferences about the data-generating process.

Both halves of \eqref{eq:demographic_parity} have the same form, so write $\Psi_{\mathrm{pDP}} = \Psi_1 - \Psi_0$ with $\Psi_g(P) = \E[D(X)\mid G=g]$.
The two components are symmetric and the influence function of a difference is the difference of the influence functions, so it suffices to treat $\Psi_1$.

There are a number of ways to compute influence functions, but a convenient and intuitive one is via point-mass contamination: $\phi_1$ is the change in the estimator with respect to an infinitesimally small perturbation of the probability distribution.
Concretely, we perturb $P$ along the point-mass contamination path
$$P_t = (1-t)P + t\,\delta_{\tilde{o}}, \qquad \tilde{o} = (\tilde{x},\tilde{g},\tilde{y}),$$
which places mass $t$ on a single observation $\tilde{o}$.
For discrete features with positive mass at $\tilde x$, the influence function is the derivative $\phi_1(\tilde{o};P) = \frac{d}{dt}\Psi_1(P_t)\big|_{t=0}$.
Appendix \ref{app:demographic-parity} carries out this calculation and verifies the result for general feature distributions using regular submodels, giving

\begin{equation}
  \phi_{\mathrm{pDP},1}(o;P) = \frac{1}{\Prob(G=1)}\Big[\pi(x)\,\big(\ind\{y=1\} - D(x)\big) + \ind\{g=1\}\big(D(x) - \Psi_1(P)\big)\Big].
  \label{eq:eif_pdp}
\end{equation}
where $\pi(x)$ is the conditional probability of positive group membership and $D(x)$ is the conditional probability of a positive outcome.

Following Section \ref{sec:setup}, we fit $\hat{D}$ and $\hat{\pi}$ on a training split, set the mean of \eqref{eq:eif_pdp} over the evaluation split to zero, and solve for $\Psi_1$:
\begin{equation}
  \bar{\Psi}_{\mathrm{pDP},1}(\hat{P}) = \frac{1}{n\,\hat{\Prob}(G=1)}\sum_{i=1}^n \Big[\hat{\pi}(x_i)\,\hat{r}_i + \ind\{g_i=1\}\,\hat{D}(x_i)\Big],
  \label{eq:est_pdp}
\end{equation}
where $\hat{r}_i = \ind\{y_i=1\} - \hat{D}(x_i)$ is the fitted residual and $\hat{\Prob}(\cdot)$ an empirical stratum probability. This expression can be readily interpreted: the second term is the average of $\hat{D}$ over the instances where $G=1$.  The first is an empirical correction, weighting errors in $\hat{D}(x_i)$ by the stratum $\hat{\pi}$ that they belong to.  Section \ref{sec:sims} demonstrates the importance of this bias correction.
Taking the difference of group-specific terms yields the full estimator:
\begin{equation}
  \begin{aligned}
    \bar{\Psi}_{\mathrm{pDP}}(\hat{P})
    &= \frac{1}{n\,\hat{\Prob}(G=1)}\sum_{i=1}^n \Big[\hat{\pi}(x_i)\,\hat{r}_i + \ind\{g_i=1\}\,\hat{D}(x_i)\Big] \\
    &\quad- \frac{1}{n\,\hat{\Prob}(G=0)}\sum_{i=1}^n \Big[(1-\hat{\pi}(x_i))\,\hat{r}_i + \ind\{g_i=0\}\,\hat{D}(x_i)\Big].
  \end{aligned}
  \label{eq:est_pdp_contrast}
\end{equation}

From Proposition \ref{prop:linear}, we can form the interval $\bar{\Psi}_{\mathrm{pDP}}(\hat{P}) \pm z_{1-\alpha/2}\,\hat{\sigma}/\sqrt{n}$, with $\hat{\sigma}^2$ the sample variance of the influence function at the fitted nuisances:
\[
  \hat{\sigma}^2 = \frac{1}{n-1} \sum_{i=1}^n \phi_{\mathrm{pDP}}(O_i;\hat{P})^2,
\]
where $\phi_{\mathrm{pDP}}=\phi_{\mathrm{pDP},1}-\phi_{\mathrm{pDP},0}$, as given in \eqref{eq:eif_pdp_contrast}.


\paragraph{Relation to the average treatment effect.}
It is worthwhile to compare this criterion to its causal analogue, an average treatment effect (ATE), since that is both a well-studied estimand for targeted learning \cite{vanderlaanTargetedLearningCausal2011} and a subject of interest in the fairness literature \cite{zhangFairnessDecisionmakingCausal2018}.
The causal analogue treats $G$ as an intervention and asks for its average effect on $Y$;
under conditional ignorability (no unobserved confounders) and positivity (all outcomes have nonzero probability) that effect is identified by the standardization, or g-formula, estimand
\[\E\big[\E[Y \mid G=1, X]\big] - \E\big[\E[Y \mid G=0, X]\big].\]
In contrast to \eqref{eq:demographic_parity} the g-formula places $G$ \emph{inside} the outcome regression and averages over the marginal law of $X$, measuring the effect of changing $G$ with $X$ held fixed.
Since group status is typically immutable, the counterfactual group membership does not have a readily defined value.
Equation \eqref{eq:demographic_parity} removes $G$ from the regression, so the rule is common to both groups, and averages over the \emph{group-conditional} law of $X$:
\[ \E \left[ \E \left[Y \mid X\right] \mid G = 1 \right] - \E \left[ \E \left[Y \mid X\right] \mid G = 0 \right] \]

\subsection{Double Robustness}
\label{sec:dr}

A nice property of this data fairness estimator for demographic parity is double robustness: the nuisance estimators for the group and outcome can compensate for each other.
Writing $\Psi_1$ for the $G=1$ component of probabilistic demographic parity, the bias of the one-step estimator factorizes as
\begin{equation}
  \E[\bar{\Psi}_1(\hat{P})\mid\mathcal T] - \Psi_1(P) = -\frac{\E\left[(\hat{\pi}(X)-\pi(X))(\hat{D}(X)-D(X))\mid\mathcal T\right]}{\Prob(G=1)} + O(n^{-1}),
  \label{eq:product}
\end{equation}
here $\mathcal T$ denotes the training data and $n$ the evaluation size.
Appendix \ref{app:proof} gives the derivation.

The product form of the two errors in \eqref{eq:product} permits a familiar double-robustness statement about the point estimate: if either nuisance is consistent, the leading bias vanishes even when the other is misspecified, and if either is exactly correct it vanishes identically.

We can also consider what this means for confidence intervals.
The product term in \eqref{eq:product} controls the remainder in Proposition \ref{prop:linear}, and by Cauchy--Schwarz its absolute value is bounded by $\|\hat{\pi}-\pi\|\,\|\hat{D}-D\|/\Prob(G=1)$.
The rate condition $R = o_P(n^{-1/2})$ is therefore met as soon as the \emph{product} of the two nuisance rates is $o_P(n^{-1/2})$, which permits each individual rate to be as slow as $o_P(n^{-1/4})$---attainable under suitable conditions on the learners and data-generating distribution \cite{kennedySemiparametricDoublyRobust2023}, as discussed in Appendix \ref{app:assume}.
We note that this property provides for the {\em consistency} of $\bar{\Psi}_1(\hat{P})$ even when one of $\hat{\pi}$ and $\hat{D}$ is not consistent, but that does not guarantee valid confidence intervals unless both are also consistently estimated.
The nuisance-misspecification simulations in Appendix~\ref{app:nuisance-simulations} illustrate that double robustness of the point estimator does not by itself guarantee nominal coverage when one nuisance model is misspecified.

\subsection{Equal Opportunity}
\label{sec:opportunity}
Parity asks whether the two groups are selected at the same rate.
Equal opportunity asks the same question of a narrower population: among those with positive outcomes, are the groups treated alike \cite{hardtEqualityOpportunitySupervised2016}?

Probabilistic equal opportunity restricts each group average to observations with positive outcomes:
\begin{equation}
  \Psi_{\mathrm{pEO}}(P) = \E[D(X) \mid Y=1,G=1] - \E[D(X) \mid Y=1,G=0].
\end{equation}
The group-1 estimator is
\begin{equation}
  \bar{\Psi}_{\mathrm{pEO},1}(\hat{P}) = \frac{1}{n\,\hat{\Prob}(Y=1,G=1)}\sum_{i=1}^n \Big[\hat{\rho}_1(x_i)\,\hat{r}_i + \ind\{y_i=1,\,g_i=1\}\,\hat{D}(x_i)\Big].
\end{equation}
Subtracting the analogous group-0 estimate gives the contrast. Appendix \ref{app:equal} provides the derivation.


\subsection{Conditional Mutual Information}
\label{sec:cmi}

The probabilistic metrics above compare average outcome probabilities across groups.
Taking a difference of means can hide associations that cancel out in the average. 
As a complementary summary of the \emph{conditional association} between $G$ and $Y$ given $X$, we therefore take the conditional mutual information (CMI) between the response and the protected attribute,
\begin{equation}
  \Psi_{\mathrm{CMI}}(P) = I(Y;G \mid X) = \E\!\left[\log \frac{p(G,Y\mid X)}{p(G\mid X)\,p(Y\mid X)}\right].
  \label{eq:cmi}
\end{equation}
CMI is non-negative and equals zero if and only if $Y \independent G \mid X$ \cite{mesnerConditionalMutualInformation2019}, and may be nonzero even when demographic parity holds.
The converse also fails: conditional independence does not specify how $X$ is distributed across groups, so a process satisfying $Y \independent G \mid X$ exactly can still show a large parity gap.

The efficient influence function of $\Psi_{\mathrm{CMI}}$ is,
\begin{equation}
  \phi(x,g,y;P) = \log \frac{p(g,y\mid x)}{p(g\mid x)\,p(y\mid x)} - \Psi_{\mathrm{CMI}}(P);
\end{equation}
See Appendix \ref{app:cmi_eif} for the derivation.
At conditional independence, $p(g,y\mid x)=p(g\mid x)\,p(y\mid x)$, so $\phi$ vanishes.
The variance of the influence function is therefore zero, so the usual Wald intervals based on this variance are not justified; the remaining estimation error is governed by higher-order terms.

To estimate CMI, we first fit a model of the joint conditional law $p(G,Y\mid X)$ over the four states of $(G,Y)$ on a training split and obtain the margins by summation.
For this initial fit $\hat P$, the one-step correction simplifies to the held-out log-ratio average,
\begin{equation}
  \hat\Psi_{\mathrm{OS}}
  =\Psi_{\mathrm{CMI}}(\hat P)+P_n\phi(O;\hat P)
  =\frac1n\sum_{i=1}^n\log\frac{\hat p(g_i,y_i\mid x_i)}{\hat p(g_i\mid x_i)\hat p(y_i\mid x_i)}.
  \label{eq:cmi_est}
\end{equation}
The one-step estimator corrects the numerical value of the initial CMI estimate using a local, linear approximation to its bias.
Because CMI is a nonlinear function of the joint probabilities, this additive adjustment can produce a negative estimate, particularly when dependence is weak.
TMLE instead uses the influence function to update the fitted joint probabilities until the empirical first-order correction is approximately zero.
Evaluating CMI at this updated distribution combines the bias-correction objective with preservation of the estimand's nonnegativity.
We therefore use TMLE for CMI estimation, with the targeting construction and conditions for inference given in Appendix \ref{app:cmi_eif}.
Writing $p^*$ for the joint law targeted using the evaluation observations, our estimate is
\begin{equation}
  \hat\Psi_{\mathrm{TMLE}}
  =\frac1n\sum_{i=1}^n\sum_{g,y}p^*(g,y\mid x_i)
  \log\frac{p^*(g,y\mid x_i)}{p_G^*(g\mid x_i)p_Y^*(y\mid x_i)}.
  \label{eq:cmi_tmle}
\end{equation}
Each inner sum is a Kullback--Leibler (KL) divergence, so this estimate is nonnegative.

We report nominal symmetric Wald intervals using the sample variance of $\hat\phi_i=\ell(O_i;P^*)-\hat\Psi_{\mathrm{TMLE}}$, where $\ell(o;P^*)$ is the log ratio of the targeted joint probability to the product of its marginals.

\subsection{Thresholded Decisions}
\label{sec:thresholded}

The fairness literature is often concerned with discrete predictions, since algorithmic predictions are in practice often translated into an action via some decision rule.
We can derive thresholded metrics to compare decision rates, replacing the score by the decision it induces,
\begin{equation}
  D_c(x) = \ind\{D(x) \geq c\}.
\end{equation}
The rule $D_c$ takes values in $\{0,1\}$: it is the action a decision-maker would take based on the conditional probability $D(x)$.
Here we treat false negatives and positives equally and take $c = \tfrac{1}{2}$ throughout.
Substituting $D_c$ for $D$ in \eqref{eq:demographic_parity} gives an estimand for thresholded demographic parity,
\begin{equation}
  \Psi_{\mathrm{DP}}(P) = \E[D_c(X)\mid G=1] - \E[D_c(X)\mid G=0].
  \label{eq:dp}
\end{equation}
Small errors in $\hat D$ can change decisions near the threshold; formally $\Psi_{DP}(P)$ does not clearly satisfy the regularity condition required for targeted learning.
For valid inference, these changed decisions must be sufficiently rare; Appendix \ref{app:threshold_stability} bounds the probability of such events.
Writing $\Psi_g(P)=\E[D_c(X)\mid G=g]$, the estimator has the same first-order expansion as if $D_c$ were known, with group-specific influence representation
\begin{equation}
  \phi_{\mathrm{DP},1}(o;P) = \frac{\ind\{g=1\}}{\Prob(G=1)}\big[D_c(x) - \Psi_1(P)\big].
  \label{eq:eif_dp}
\end{equation}
The estimator is a bare plug-in, the average of the fitted decision over the appropriate strata of $G$,
\begin{equation}
  \bar{\Psi}_{\mathrm{DP}}(\hat{P}) = \frac{1}{n\,\hat{\Prob}(G=1)}\sum_{i=1}^n \ind\{g_i=1\}\,\hat{D}_c(x_i) - \frac{1}{n\,\hat{\Prob}(G=0)}\sum_{i=1}^n \ind\{g_i=0\}\,\hat{D}_c(x_i),
  \label{eq:est_dp}
\end{equation}
Thresholded equal opportunity compares decision rates among observations with positive outcomes,
\begin{equation}
  \Psi_{\mathrm{EO}}(P) = \E[D_c(X)\mid Y=1,G=1] - \E[D_c(X)\mid Y=1,G=0].
\end{equation}
Its group-1 estimator is
\begin{equation}
  \bar{\Psi}_{\mathrm{EO},1}(\hat{P}) = \frac{1}{n\,\hat{\Prob}(Y=1,G=1)}\sum_{i=1}^n \ind\{y_i=1,\,g_i=1\}\,\hat{D}_c(x_i).
\end{equation}
Subtracting the analogous group-0 estimate gives the contrast. Appendix \ref{app:threshold_stability} derives its influence representation and the stability conditions required for inference.

Because thresholding is discontinuous, small estimation errors can flip decisions near $c$; the margin condition in Appendix \ref{app:threshold_stability} limits the probability mass near this boundary so that, with sufficiently accurate estimation of $D$, these flips are negligible relative to sampling uncertainty.

These assumptions also obviate the need for an influence-function correction: estimation errors affect the thresholded metric only through changed decisions, whose contribution is negligible at first order.
The thresholded data-fairness estimators and their Wald intervals therefore coincide with those for model fairness evaluated using the same fitted rule and evaluation sample, although the estimands differ.
These assumptions are demanding and may fail in practice, in which case the intervals are not guaranteed to cover the population data-fairness estimand.
Smooth approximations to the threshold, with appropriate control of approximation error, offer a possible extension that we leave to future work.

\section{Simulations}
\label{sec:sims}

To evaluate the properties of our data fairness estimators, we perform several simulations.
First, we compare and contrast the coverage and power of inference on data fairness with inference on model fairness across independently sampled training and evaluation datasets.
Next, we investigate how sensitive our estimators are to the accuracy of nuisance estimation, varying the amount of training data and model capacity.
Lastly, we demonstrate how data fairness inference can be used to evaluate the choice of a model feature set.

\subsection{Coverage Under Independent Retraining}
\label{sec:retraining}

We investigate how variation in the training data affects inference on data fairness.
We draw $X=(U,V)$ uniformly from $\{-1,1\}^2$, and draw $G$ and $Y$ independently given $X$, with
\[
  \Prob(G=1\mid X)=(1+0.6U)/2, \qquad \Prob(Y=1\mid X)=\expit(V+\delta U),
\]
where $\expit(u)=1/(1+e^{-u})$.
We vary $\delta$ over $\{0, 0.2, 0.6\}$, representing a population null and two alternatives with increasing parity disparities.
When $\delta=0$, a fitted model may learn a spurious relationship between $U$ and $Y$, producing a disparity absent from the population regression. We consider these false positives.

We vary training size over nine roughly log-spaced values from $100$ to $2000$, and evaluation size over $\{250, 500, 1000, \allowbreak 2000, 4000\}$.
For each configuration, we draw $1000$ independent training datasets and fit the outcome and propensity models using the observed proportions in the four feature cells. 
For each fitted model, we draw $50$ independent evaluation datasets.
We compare TL with a difference in mean predictions between groups, treating the fitted model as fixed.
Both estimators use the same observations, although TL additionally uses the evaluation outcomes for its residual correction.
We measure coverage, rejection rates, and interval widths; Monte Carlo standard errors account for evaluations sharing a fitted model.

\begin{figure}[t]
  \centering
  \includegraphics[page=1,width=\linewidth]{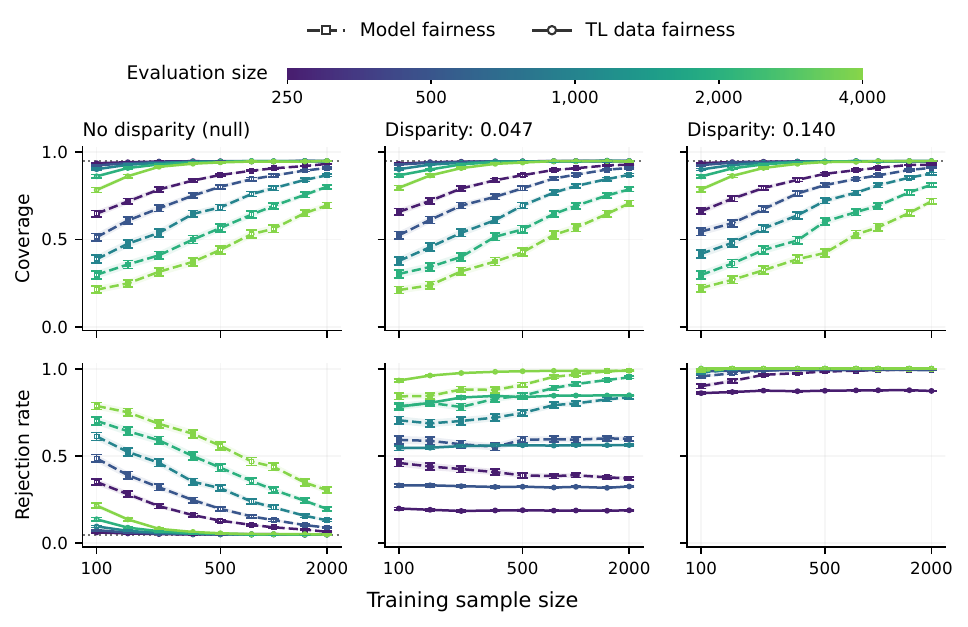}
  \caption{Coverage (top) and rejection rate (bottom) under independent retraining. Columns show the null and two population disparities. Color indicates evaluation size; line style distinguishes TL and model fairness. Reference lines mark nominal coverage $0.95$ and null rejection rate $0.05$. Under the alternatives, rejection is power. Each point uses 1000 independent training datasets and 50 evaluations per fit; bars and bands show $\pm1.96$ Monte Carlo standard errors accounting for shared fits.}
  \label{fig:retraining}
\end{figure}

Figure~\ref{fig:retraining} shows that TL improves coverage of the population parity gap across all configurations in this experiment, although both sample sizes matter.
Under the null, increasing evaluation size while holding training size fixed can worsen model-based coverage of the population disparity, while TL remains closer to nominal coverage.
Model-based intervals remain approximately calibrated for the fitted model's own disparity, illustrating the distinction between inference on model fairness and data fairness.


\subsection{Training Size and Nuisance Estimation}
\label{sec:trainsize}
\label{sec:robustness}

Figure~\ref{fig:retraining} shows that training variability affects inferences about population fairness, and that model fairness intervals can badly undercover the true population disparity.
A natural question is whether this distinction becomes negligible with more training data and a flexible model: if the fitted outcome probabilities are sufficiently accurate, how does data fairness still offer an advantage over model fairness?
To examine this we vary training size and learner capacity, comparing both methods against the same population data-fairness target.

We extend the shared-predictor structure of Figure~\ref{fig:retraining} to continuous, standard-normal features and nonlinear relationships.
One predictor drives both group membership and the outcome, another predicts only the outcome, and additional features contain no signal.
This creates a learning problem in which both limited training data and approximation error can affect the fitted probabilities.
We compare lower- and higher-capacity gradient-boosted decision-tree classifiers, both with early stopping, as training size increases while evaluation size stays fixed.
Within each comparison, TL and model fairness use the same outcome predictions and observations, so their difference isolates what TL's residual correction adds.
Greater capacity allows more complex fits but does not guarantee greater accuracy, especially with limited data.

\begin{figure}[t]
  \centering
  \includegraphics[width=\linewidth]{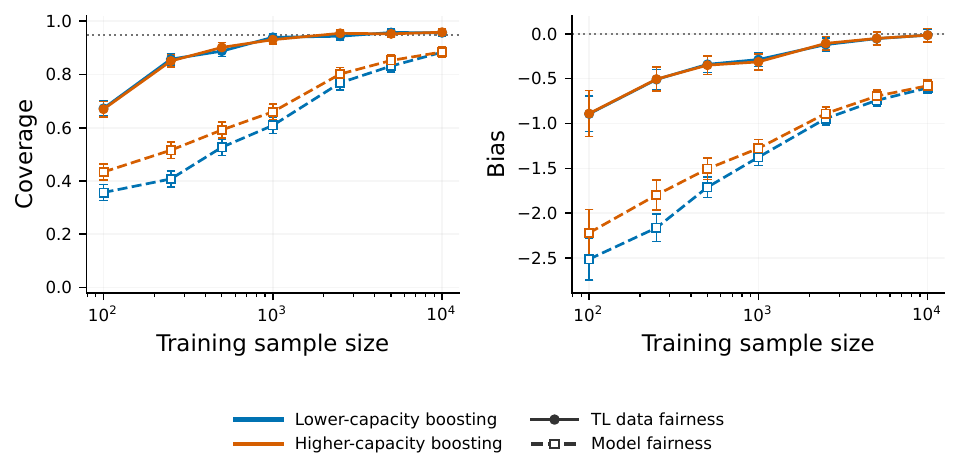}
  \caption{Coverage (left) and signed bias relative to population data fairness (right) as training size increases. Bias is $100\{R^{-1}\sum_{r=1}^{R}\hat\Psi_r-\Psi\}$ in percentage points. Each point uses $R=1000$ replicates and 2000 evaluation observations. Color identifies tree-ensemble capacity; solid circles denote TL and dashed squares denote model fairness. Bars show approximate pointwise Monte Carlo uncertainty ($\pm1.96$ MCSE), not the fairness intervals evaluated within replicates. Reference lines mark $0.95$ coverage and zero bias.}
  \label{fig:trainsize}
\end{figure}

Figure~\ref{fig:trainsize} shows that more training data improves inference, but does not quickly eliminate the distinction between model and data fairness.
Both fitted learners understate the population gap on average; TL substantially reduces this bias and approaches nominal coverage sooner.
The two capacity settings give similar TL results, and neither achieves nominal coverage at the smallest training sizes.
Appendix~\ref{app:nuisance-simulations} gives the simulation details and compares true conditional probabilities with correctly specified and misspecified fitted models to distinguish bias correction from confidence-interval calibration.

\subsection{Selecting a Feature Set}
\label{sec:conditioning}

An advantage of thinking about fairness in the data-generating process is that it shifts the focus from model selection to feature selection.
Specifically, we can reason about which features to include in a model by evaluating conditional mutual information (CMI) between the outcome and group membership given a candidate feature set.

\begin{figure}
  \centering
  \includegraphics[width=0.9\linewidth]{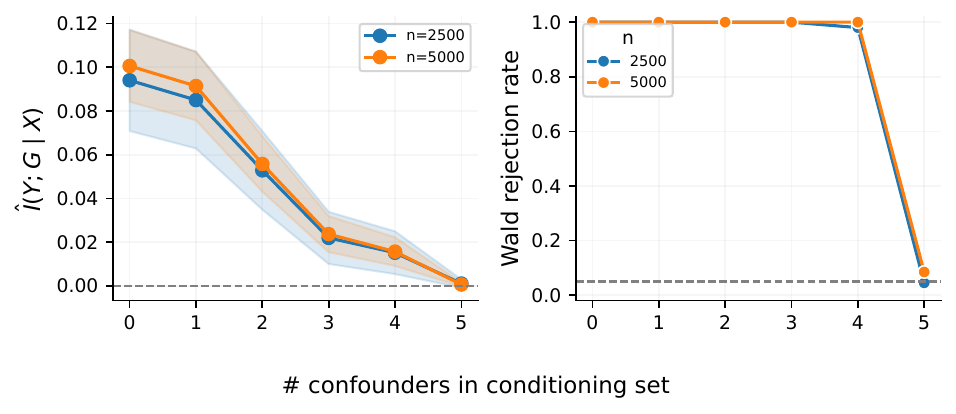}
  \caption{CMI estimates and empirical Wald rejection rates as the conditioning set changes. The true CMI is zero when all confounders $Z$ are included; outcome-only predictors and noise features are always retained. Left: mean TMLE estimate, with shading between the mean endpoints of the nominal $95\%$ two-sided confidence intervals $\hat I\pm1.96\,\widehat{\mathrm{se}}$. Right: empirical rejection rate of the one-sided rule $\hat I-1.645\,\widehat{\mathrm{se}}>0$, with the dashed line at the nominal $0.05$ level. Each point uses 200 simulations.}
  \label{fig:condset}
\end{figure}

The outcome and group are conditionally independent given the features $X$ if $I(Y;G\mid X)=0$.
Here we simulate a confounded data-generating process and vary which features are conditioned on, to show how the estimate depends on that choice.
We draw $p$ independent standard-normal features partitioned into three groups: confounders $Z$ that drive both $G$ and $Y$, outcome-only predictors $X_O$ that drive $Y$ alone, and noise.
We generate $G \sim \Bernoulli(\expit(Z\alpha))$ and $Y \sim \Bernoulli(\expit(Z\beta_c + X_O\beta_o))$, drawing $Y$ and $G$ independently given the features.
By construction $Y \independent G \mid Z$ holds exactly, so the true CMI is zero for any conditioning set that contains $Z$; yet $Y$ and $G$ remain \emph{marginally} dependent through the shared confounders (we confirm a nonzero empirical $I(Y;G)$).
We sweep $k$, the number of confounders included in the conditioning set, always retaining the outcome predictors and noise.
We also report the empirical rejection rate of the one-sided Wald rule, which rejects $H_0:I(Y;G\mid X)=0$ when the nominal $95\%$ lower confidence bound $\hat{I}(Y;G\mid X) - 1.645\,\widehat{\mathrm{se}}$ exceeds zero, where $\widehat{\mathrm{se}}$ is the estimated standard error of $\hat I$.
We use TMLE for the CMI estimates. Figure \ref{fig:condset} displays nominal two-sided confidence intervals alongside empirical rejection rates for the one-sided rule. The nominal $0.05$ cutoff does not establish calibration at conditional independence. Appendix \ref{app:cmi} compares one-step and TMLE estimates computed from identical initial probability estimates and evaluation observations, isolating the effect of targeting on estimates and interval calibration.

Figure \ref{fig:condset} shows that the estimates and Wald rejection rates change with the conditioning set.
With all confounders included, the true CMI is zero by construction, providing a direct check of the Wald rule at the degenerate null.
The targeted estimates are nonnegative, while their nominal lower confidence bounds may be negative.
Low rejection rates do not establish conditional independence or, by themselves, validate interval coverage.
Adjusting for features that predict $Y$ well is therefore not the same as adjusting for the features that render $Y$ and $G$ conditionally independent.

\section{Data Analysis}
\label{sec:data}

We show how our approach to data fairness can be used in practice by applying our estimators to real data.
We analyze the Adult-Income dataset and the Law School dataset, two datasets common in the fairness literature, and discuss the interpretations and implications of our results from a data fairness perspective.
To perform our analysis, we split the data into a 60/40 train/test split. For parity and equal opportunity, we train SuperLearner ensembles \cite{laanSuperLearner2007} for the outcome, group, and joint group--outcome distributions. For CMI, we fit a gradient-boosted classifier to the joint group--outcome distribution, using predictive performance on a validation subset of the training data to determine when to stop training (Appendix~\ref{app:cmi}). We then target this law as described in Section \ref{sec:cmi} and derive its marginals from the targeted fit.
Appendix~\ref{app:data-details} specifies the dataset versions, features, preprocessing, and ensemble fitting procedure.
For both datasets, our analysis seeks to determine to what extent the distribution of $Y$ given $X$ violates definitions of fairness.
We also report model-fairness intervals conditional on the fitted outcome model. Both targets use the same outcome predictions and held-out observations. The probabilistic data-fairness estimators include the nuisance corrections described above, whereas the model-fairness estimators compare group means of the fitted scores. With the same empirical influence-function variance convention, the thresholded estimates and intervals coincide numerically for the two targets.
Thresholded inference assumes the stability conditions in Appendix \ref{app:threshold_stability}, which we have not verified for these fits.

\begin{table}
  \centering
  \small
\begin{tabular}{@{}llcc@{}}
\toprule
Metric & Dataset & Data fairness & Model fairness \\
\midrule
Prob. Parity & Adult & 0.180 (0.170, 0.191) & 0.178 (0.170, 0.185) \\
 & Law & 0.192 (0.172, 0.212) & 0.194 (0.179, 0.208) \\
\addlinespace
Prob. Eq. Opp. & Adult & 0.072 (0.039, 0.105) & 0.101 (0.074, 0.129) \\
 & Law & 0.135 (0.120, 0.150) & 0.127 (0.113, 0.141) \\
\midrule
Parity (thresholded) & Adult & 0.176 (0.166, 0.186) & 0.176 (0.166, 0.186) \\
 & Law & 0.228 (0.204, 0.252) & 0.228 (0.204, 0.252) \\
\addlinespace
Eq. Opp. (thresholded) & Adult & 0.122 (0.082, 0.162) & 0.122 (0.082, 0.162) \\
 & Law & 0.122 (0.100, 0.144) & 0.122 (0.100, 0.144) \\
\midrule
CMI (TMLE) & Adult & 0.0008 (0.0003, 0.0014) & -- \\
 & Law & 0.0001 (-0.0002, 0.0005) & -- \\
\addlinespace
\bottomrule
\end{tabular}

  \caption{Data- and model-fairness estimates with nominal $95\%$ confidence intervals in parentheses. Parity and equal-opportunity comparisons share the fitted outcome model and evaluation split. CMI uses TMLE (Section \ref{sec:cmi}) and is reported only for data fairness.}
  \label{tab:data_results}
\end{table}

\paragraph{Adult-Income.}
We use the IPUMS reconstruction of Adult-Income released by Ding et al.\ \cite{dingRetiringAdultNew2021}, containing economic and demographic information on individuals living in the United States \cite{floodIntegratedPublicUse2020}.
For our analysis, the outcome is $Y = \ind\{\text{income} > 50,000\}$ and $G = \ind\{\text{is male}\}$.
We find that both parity and probabilistic parity are violated, with confidence intervals bounded away from zero.
The CMI estimate is small, and its confidence interval excludes zero.

\paragraph{Law School.}
We use the cleaned Law School dataset distributed with Le Quy et al.'s survey \cite{lequySurveyDatasetsFairnessaware2022}, containing educational, demographic, and economic information from the LSAC study of law students \cite{wightmanLSACNationalLongitudinal1998}.
Here, our outcome of interest is $Y=\ind\{\text{passed the bar}\}$ and $G = \ind\{\text{is white}\}$.
Again we obtain positive estimates with intervals bounded away from zero for parity and equal opportunity.
We note that both equal opportunity estimates have narrower intervals than in the Adult dataset.
Positive outcomes are less prevalent in Adult than in Law, so conditioning on $Y=1$ retains a smaller fraction of the Adult observations.
This is consistent with the wider equal-opportunity intervals observed for Adult.
The CMI estimate is also small, with a confidence interval that includes zero.

\section{Discussion}
\label{sec:discussion}

In this paper, we define and advocate for data fairness as a target of inference.
Our thesis is that one should think about fairness in algorithmic decision-making as a property of the underlying data-generating process, rather than as a property of a particular model trained on that data.
In motivating this perspective, we distinguish between multiple sources of uncertainty that affect the conclusions one can draw about fairness.
In order to perform inference on unknown data distributions, we apply targeted learning to construct estimators for data fairness metrics.
Simulations demonstrate that fixed-model confidence intervals can under-cover data fairness severely, while estimating data fairness with targeted learning improves coverage substantially.


\paragraph{Limitations and Future Work}

Our work has several limitations that connect with potential future work.
First, the fairness metrics we consider (demographic parity and equal opportunity) both operate on binary group status and binary outcomes.
Although binary metrics are common in algorithmic fairness, focusing only on binary group status is limiting.
In the cases of gender, wealth or age discrimination, for example, one would be interested in evaluating fairness on categorical or continuous attributes.
Further work should develop tools for inference on data fairness that allow categorical and continuous attributes and outcomes.

Additionally, CMI inference requires care at conditional independence, where the efficient influence function vanishes. Targeting preserves nonnegative point estimates, but does not supply a nondegenerate first-order Wald approximation at this null.
Further work should explore inference in this setting, particularly considering the tradeoffs between binary assessments of evidence and estimation.

Our work should not be interpreted as an argument that model fairness is unimportant, or that data fairness is necessarily a replacement for model fairness.
As noted in the introduction, a model trained on a data-fair process may still be unfair if trained on a small or unrepresentative sample of the data.
Such a model should not be deployed; rather, data fairness is a tool to help identify whether retraining on a larger or more representative sample is likely to produce a more fair model.
The distinction between a bad sample and a bad data process is a crucial one for institutions developing their data collection and model training pipelines.
Ultimately we view data fairness as a complement to model fairness, and one of our main contributions is to provide language and tools to differentiate between the multiple sources of uncertainty that can affect fairness in algorithmic decision-making.

\clearpage
\section*{Generative AI Usage Disclosure}
We used OpenAI Codex with models ranging from GPT-5.5 to GPT-6 Astra and Claude Code with models ranging from Claude Opus 4.8 to Claude Opus 5.5.
Generative AI assistance included drafting and revising manuscript text, identifying potentially relevant literature, exploring and checking statistical arguments and mathematical derivations, developing and debugging analysis code, running simulations and real-data analyses, and revising figures.
We also used Consensus to assist with the literature review.
References were managed with Zotero rather than generated by AI, to guard against fabricated sources.
This work was directed by the authors, including iterative review and revision of proposed manuscript passages and discussion of statistical arguments.
The authors retain full and non-delegable responsibility for the correctness, originality, integrity, and presentation of all content in this work, including AI-assisted text, code, analyses, and figures, as well as all references.

\bibliographystyle{ACM-Reference-Format}
\bibliography{refs}

\appendix

\section{Targeted Learning Requirements for Probabilistic Metrics}
\label{app:assume}

This appendix details the sample-splitting and nuisance-estimation conditions for probabilistic demographic parity and equal opportunity.
The separate argument for thresholded metrics is given in Appendix \ref{app:threshold_stability}, and the CMI derivation and its degeneracy at conditional independence are discussed in Appendix \ref{app:cmi_eif}.

As discussed in Section \ref{sec:inference}, the error in our estimate can be shown to be
\begin{align*}
  \Psi(\hat{P}) - \Psi(P)
  &= P_n[\phi(O;P)] - P_n[\phi(O;\hat{P})] + (P_n - P)[\phi(O;\hat{P}) - \phi(O;P)] + R(\hat{P},P),
\end{align*}
where $P_n$ is the empirical distribution of the evaluation sample. 
After adding back the plug-in bias as in Section \ref{sec:setup}, the corrected estimator $\bar{\Psi}(\hat{P})$ gives:
\begin{align*}
  \bar{\Psi}(\hat{P}) - \Psi(P)
  &= P_n[\phi(O;P)] + (P_n - P)[\phi(O;\hat{P}) - \phi(O;P)] + R(\hat{P},P),
\end{align*}

To use the central limit theorem for inference, we require that $\bar{\Psi}(\hat{P}) - \Psi(P) = P_n[\phi(O;P)] + o_P(n^{-1/2})$.
This means the empirical process and remainder terms must each be $o_P(n^{-1/2})$.


To control the empirical process term $(P_n - P)[\phi(O;\hat{P}) - \phi(O;P)]$, we fit $\hat{P}$ on a training sample independent of the $n$ evaluation observations used to form $P_n$.
Conditional on the training data $\mathcal T$, the influence-function difference $\phi(\,\cdot\,;\hat{P})-\phi(\,\cdot\,;P)$ is a fixed function, and the evaluation observations remain independent draws from $P$.
Thus the empirical process term is an average of conditionally independent, mean-zero variables, with conditional variance
\[
  \begin{aligned}
    &\Var\!\left((P_n-P)[\phi(O;\hat{P})-\phi(O;P)]\mid\mathcal T\right) \\
    &\qquad =\frac{\Var_P[\phi(O;\hat{P})-\phi(O;P)]}{n}
    \leq\frac{\|\phi(\,\cdot\,;\hat{P})-\phi(\,\cdot\,;P)\|^2}{n},
  \end{aligned}
\]
where $\|\cdot\|$ is the $\mathcal L_2(P)$ norm used in Section \ref{sec:setup}.
If the influence function is $\mathcal L_2$-consistent, $\|\phi(\,\cdot\,;\hat{P})-\phi(\,\cdot\,;P)\|=o_P(1)$, the variance is $o_P(n^{-1})$.
Plugging this variance into Chebyshev's inequality gives, for any $\epsilon>0$,
\[
  \begin{aligned}
    &\Prob\!\left(
      \left|(P_n-P)[\phi(O;\hat{P})-\phi(O;P)]\right|
      >\frac{\epsilon}{\sqrt n}\,\middle|\,\mathcal T
    \right) \leq
    \frac{\|\phi(\,\cdot\,;\hat{P})-\phi(\,\cdot\,;P)\|^2/n}{\epsilon^2/n}
    =o_P(1).
  \end{aligned}
\]
Sample splitting and $\mathcal{L}_2$ consistency therefore supply the desired rate \cite{kennedySemiparametricDoublyRobust2023}.

We also require that the remainder term $R(\hat{P},P)$ be $o_P(n^{-1/2})$.
This condition demands additional control of nuisance-estimation error.
For probabilistic demographic parity, the product-bias representation in Section \ref{sec:dr} and Appendix \ref{app:proof} gives the sufficient condition
\[
  \|\hat D-D\|\,\|\hat\pi-\pi\|=o_P(n^{-1/2}).
\]
Both nuisances converging at $o_P(n^{-1/4})$ suffices, but faster convergence of one permits slower convergence of the other.
A practical strategy for improving nuisance-estimation rates is to fit ensembles of learners with different modeling assumptions, increasing the chance that some candidates capture the relevant structure in $D$, $\pi$, or $\rho_g$ (for equal opportunity).
Super Learner combines candidate learners using cross-validated predictions \cite{laanSuperLearner2007}. Such combinations can improve estimation accuracy, although attaining the required product rate still depends on the candidate learners and the data-generating distribution.
Thresholded metrics instead require the decision-boundary control developed in Appendix \ref{app:threshold_stability}.

Finally, we assume mild positivity conditions: there must be nonzero probability for each stratum of our data. Concretely we assume $D(x),\pi(x)\in(0,1)$ and $\Prob(G=g)$ is bounded away from zero for both groups; equal opportunity additionally requires $\Prob(Y=1,G=g)$ to be bounded away from zero.

\section{Probabilistic Demographic Parity EIF Derivation}
\label{app:demographic-parity}

Recall that $D(x)=\Prob(Y=1\mid X=x)$ and $\pi(x)=\Prob(G=1\mid X=x)$.
Write $\Psi_{\mathrm{pDP}}(P)=\Psi_1(P)-\Psi_0(P)$, where $\Psi_g(P)=\E[D(X)\mid G=g]$.
For the group-1 component, define the numerator and denominator explicitly:
\[
  \Psi_1(P)=\frac{a}{b},\qquad
  a=\E[\ind\{G=1\}D(X)],\qquad b=\Prob(G=1)
\]
Perturb the joint distribution of $O=(X,G,Y)$ along
\[
  P_t=(1-t)P+t\delta_{\tilde o},\qquad
  \tilde o=(\tilde x,\tilde g,\tilde y).
\]
Let $D_t(x)=\Prob_{P_t}(Y=1\mid X=x)$, $a_t=\E_{P_t}[\ind\{G=1\}D_t(X)]$, and $b_t=\Prob_{P_t}(G=1)$.
A dot denotes differentiation at $t=0$.
For the point-mass calculation, first take $X$ to be discrete, with $f_X(\tilde x)=\Prob(X=\tilde x)>0$; we verify the result for continuous features using regular submodels below.

Both the joint probability of $(X=x,Y=1)$ and the marginal probability of $X=x$ change under contamination, so
\[
  D_t(x)=
  \frac{(1-t)f_X(x)D(x)+t\ind\{x=\tilde x\}\ind\{\tilde y=1\}}
  {(1-t)f_X(x)+t\ind\{x=\tilde x\}}.
\]
Applying the quotient rule, for $f_X(x)>0$,
\begin{align*}
  \dot D(x)
  &=\frac{[\ind\{x=\tilde x\}\ind\{\tilde y=1\}-f_X(x)D(x)]f_X(x)
  -f_X(x)D(x)[\ind\{x=\tilde x\}-f_X(x)]}{f_X(x)^2} \\
  &=\frac{\ind\{x=\tilde x\}}{f_X(x)}[\ind\{\tilde y=1\}-D(x)].
\end{align*}
The terms involving $f_X(x)^2D(x)$ cancel: only the outcome residual at the contaminating feature value remains.

The denominator is
\[
  b_t=(1-t)b+t\ind\{\tilde g=1\},\qquad
  \dot b=\ind\{\tilde g=1\}-b.
\]
The numerator changes in two ways: the distribution being averaged over changes, and the function $D_t$ being averaged changes.
Writing out the mixture before differentiating makes both contributions explicit:
\begin{align*}
  a_t&=(1-t)\E_P[\ind\{G=1\}D_t(X)]
  +t\ind\{\tilde g=1\}D_t(\tilde x), \\
  \dot a&=-a+\E_P[\ind\{G=1\}\dot D(X)]
  +\ind\{\tilde g=1\}D(\tilde x).
\end{align*}
Conditioning on $X$ replaces $\ind\{G=1\}$ in the middle expectation by $\pi(X)$, giving
\begin{equation}
  \dot a=\ind\{\tilde g=1\}D(\tilde x)-a
  +\E_P[\pi(X)\dot D(X)].
  \label{eq:adot}
\end{equation}
Substituting the expression for $\dot D$,
\begin{align*}
  \E_P[\pi(X)\dot D(X)]
  &=\sum_x f_X(x)\pi(x)
  \frac{\ind\{x=\tilde x\}}{f_X(x)}[\ind\{\tilde y=1\}-D(x)] \\
  &=\pi(\tilde x)[\ind\{\tilde y=1\}-D(\tilde x)].
\end{align*}
This is the contribution from estimating the outcome probability itself; it would be absent if $D$ were replaced by a fixed model prediction.

Differentiating $\Psi_1(P_t)=a_t/b_t$ gives
\begin{equation}
  \phi_{\mathrm{pDP},1}(\tilde o;P)
  =\left.\frac{d}{dt}\Psi_1(P_t)\right|_{t=0}
  =\frac{\dot a\,b-a\,\dot b}{b^2}
  =\frac{\dot a-\Psi_1(P)\dot b}{b}.
  \label{eq:ratio}
\end{equation}
The numerator simplifies to
\begin{align*}
  \dot a-\Psi_1(P)\dot b
  &=\ind\{\tilde g=1\}D(\tilde x)-a
  +\pi(\tilde x)[\ind\{\tilde y=1\}-D(\tilde x)] \\
  &\quad-\Psi_1(P)[\ind\{\tilde g=1\}-b] \\
  &=\pi(\tilde x)[\ind\{\tilde y=1\}-D(\tilde x)]
  +\ind\{\tilde g=1\}[D(\tilde x)-\Psi_1(P)],
\end{align*}
where $-a+\Psi_1(P)b=0$.
Define the outcome residual
\begin{equation}
  r(o)=\ind\{y=1\}-D(x).
  \label{eq:residual}
\end{equation}
Relabelling $\tilde o$ as $o$ recovers \eqref{eq:eif_pdp}:
\[
  \phi_{\mathrm{pDP},1}(o;P)=\frac{1}{\Prob(G=1)}
  \Big[\pi(x)r(o)+\ind\{g=1\}[D(x)-\Psi_1(P)]\Big].
\]
The first term accounts for variation in the outcome probability; the second accounts for variation in its group-specific average.
Both have mean zero: $\E[r(O)\mid X]=0$, and
$\E[\ind\{G=1\}(D(X)-\Psi_1(P))]=a-b\Psi_1(P)=0$.
Since $D$, $\pi$, and $r$ are bounded and $b>0$, the influence function is also square-integrable.

For continuous $X$, a literal point mass need not define a regular perturbation of $P$.
Instead, consider a regular submodel through $P$ with score $s(O)$, the derivative of its log density at $t=0$, satisfying $P[s(O)]=0$.
Differentiating the conditional mean by the quotient rule gives
\[
  \dot D(x)=\E[\ind\{Y=1\}s(O)\mid X=x]-D(x)\E[s(O)\mid X=x]
  =\E[r(O)s(O)\mid X=x].
\]
The product rule for the numerator and differentiation of the denominator yield
\begin{align*}
  \dot a
  &=P[\ind\{G=1\}D(X)s(O)]+P[\pi(X)\dot D(X)] \\
  &=P\!\left[\{\ind\{G=1\}D(X)+\pi(X)r(O)\}s(O)\right], \\
  \dot b&=P[\ind\{G=1\}s(O)].
\end{align*}
Thus \eqref{eq:ratio} gives $\left.\frac{d}{dt}\Psi_1(P_t)\right|_{t=0}=P[\phi_{\mathrm{pDP},1}(O;P)s(O)]$.
This identity, together with mean-zero and square-integrability, verifies the efficient influence function in the unrestricted nonparametric model \cite{hinesDemystifyingStatisticalLearning2022}.

For group 0, replace $\pi(x)$ by $1-\pi(x)$, the group indicator by $\ind\{g=0\}$, and the denominator by $\Prob(G=0)$.
Subtracting the two components gives
\begin{equation}
  \begin{aligned}
    \phi_{\mathrm{pDP}}(o;P)
    &=\left[\frac{\pi(x)}{\Prob(G=1)}-
    \frac{1-\pi(x)}{\Prob(G=0)}\right]r(o) \\
    &\quad+\frac{\ind\{g=1\}}{\Prob(G=1)}[D(x)-\Psi_1(P)]
    -\frac{\ind\{g=0\}}{\Prob(G=0)}[D(x)-\Psi_0(P)].
  \end{aligned}
  \label{eq:eif_pdp_contrast}
\end{equation}
With nuisances fitted on an independent training sample, write
$\hat r_i=\ind\{y_i=1\}-\hat D(x_i)$ and
$\hat{\Prob}(G=g)=n^{-1}\sum_i\ind\{g_i=g\}$.
For group 1, setting the empirical influence-function mean to zero gives
\[
  0=\sum_{i=1}^n\Big[\hat\pi(x_i)\hat r_i
  +\ind\{g_i=1\}\{\hat D(x_i)-\bar{\Psi}_{\mathrm{pDP},1}(\hat{P})\}\Big].
\]
Moving the term containing $\bar{\Psi}_{\mathrm{pDP},1}(\hat{P})$ to the other side and dividing by the group count yields
\[
  \bar{\Psi}_{\mathrm{pDP},1}(\hat{P})=\frac{\sum_i[\hat\pi(x_i)\hat r_i
  +\ind\{g_i=1\}\hat D(x_i)]}
  {n\hat{\Prob}(G=1)}.
\]
The group-0 equation uses $1-\hat\pi(x_i)$ in its residual correction.

\section{Double Robustness Proof}
\label{app:proof}

We derive the product-bias representation~\eqref{eq:product} for the $G=1$
component $\bar{\Psi}_1(\hat{P})=\bar{\Psi}_{\mathrm{pDP},1}(\hat{P})$ of the probabilistic demographic parity estimator. The
$G=0$ component is identical, and probabilistic equal opportunity follows on
replacing $\ind\{G=1\}$ by $\ind\{Y=1,G=1\}$, $\pi$ by $\rho_1$, and $\Prob(G=1)$ by
$\Prob(Y=1,G=1)$ throughout.

Condition on the training data $\mathcal{T}$, so that $\hat\pi$ and $\hat D$
are fixed functions taking values in $[0,1]$. As in Section \ref{sec:setup},
$P[f(O)]$ denotes expectation over a fresh observation and
$P_n[f(O)]=n^{-1}\sum_{i=1}^n f(O_i)$ the average over the independent evaluation sample.
Write $S_1(O)=\ind\{G=1\}$ and define
\[
  p_1=\Prob(G=1)=P[S_1(O)]>0,\qquad
  \hat p_1=\hat{\Prob}(G=1)=P_n[S_1(O)].
\]
Recall that $\Psi_1(P)=P[\pi(X)D(X)]/p_1$, and write
$\hat r(O)=\ind\{Y=1\}-\hat D(X)$, so $\hat r(O_i)=\hat r_i$ as in Section \ref{sec:parity}.
The estimator is
\begin{equation}
\label{eq:dr-A}
  \bar{\Psi}_1(\hat{P}) \;=\; \frac{P_n[A]}{\hat p_1},
  \qquad
  A \;=\; \hat\pi(X)\hat r(O)+S_1(O)\,\hat D(X).
\end{equation}
Since $\E[\ind\{Y=1\}\mid X]=D(X)$ and $\E[S_1(O)\mid X]=\pi(X)$ some manipulation provides
\begin{equation}
\label{eq:dr-PA}
  \begin{aligned}
    P[A]&=P\bigl[\hat\pi(X)\{D(X)-\hat D(X)\}+\pi(X)\hat D(X)\bigr] \\
        &=P[\pi(X)D(X)]-C, \\
    C&=P\bigl[\{\hat\pi(X)-\pi(X)\}\{\hat D(X)-D(X)\}\bigr].
  \end{aligned}
\end{equation}
Let $\mu=P[A]/p_1=\Psi_1(P)-C/p_1$ and $U=P_n[A-\mu S_1(O)]$. Then $U$ is an average of
i.i.d.\ bounded terms with mean $P[A]-\mu p_1=0$, and whenever $\hat p_1>0$,
\begin{equation}
\label{eq:dr-exact}
  \bar{\Psi}_1(\hat{P})-\Psi_1(P)
  \;=\; \frac{P_n[A]-\mu\,P_n[S_1(O)]}{\hat p_1}+\mu-\Psi_1(P)
  \;=\; -\frac{C}{p_1}+\frac{U}{\hat p_1}.
\end{equation}
This identity is exact: the error is the product term $-C/p_1$ plus a mean-zero
average divided by $\hat p_1$. Everything that follows only replaces $1/\hat p_1$
by $1/p_1$ in the second term, using
\begin{equation}
\label{eq:dr-ratio}
  \frac{U}{\hat p_1} \;=\; \frac{U}{p_1}-\frac{U(\hat p_1-p_1)}{p_1\,\hat p_1},
  \qquad
  \E\bigl[U^2\mid\mathcal{T}\bigr]=O(n^{-1}),
  \qquad
  \E\bigl[(\hat p_1-p_1)^2\bigr]=\frac{p_1(1-p_1)}{n}.
\end{equation}

For the conditional bias, set $\bar{\Psi}_1(\hat{P})=0$ when $\hat p_1=0$. On the event
$\{\hat p_1\ge p_1/2\}$ the last term of~\eqref{eq:dr-ratio} is at most
$2\,|U|\,|\hat p_1-p_1|/p_1^2$ in absolute value, which has expectation $O(n^{-1})$
by Cauchy--Schwarz. The complementary event has probability at most
$e^{-np_1^2/2}$ by Hoeffding's inequality, and on it
$|\bar{\Psi}_1(\hat{P})-\mu|$ is at most a constant multiple of $n$, so its contribution
is $o(n^{-1})$. Since $\E[U\mid\mathcal{T}]=0$,
\begin{equation}
\label{eq:dr-bias-proof}
  \E\bigl[\bar{\Psi}_1(\hat{P})\mid\mathcal{T}\bigr]-\Psi_1(P)
  \;=\; -\frac{C}{p_1}+O(n^{-1}),
\end{equation}
uniformly over the fitted nuisances. 

Double robustness is now immediate. If $\hat\pi=\pi$ or $\hat D=D$ then
$C=0$, however badly the other nuisance is misspecified. In general, by
Cauchy--Schwarz,
\[
  |C| \;\le\; \|\hat\pi-\pi\|\,\|\hat D-D\|,
\]
so $C=o_P(1)$ as soon as either nuisance is $\mathcal{L}_2$-consistent, and
$\bar{\Psi}_1(\hat{P})$ is then consistent by~\eqref{eq:dr-exact}. Moreover
$C=o_P(n^{-1/2})$ as soon as the product of the two convergence rates is,
which holds in particular if each nuisance converges at rate $o_P(n^{-1/4})$. 

For inference, \eqref{eq:dr-exact} and~\eqref{eq:dr-ratio} give
\begin{equation}
\label{eq:dr-linear}
  \bar{\Psi}_1(\hat{P})-\Psi_1(P)
  \;=\; -\frac{C}{p_1}+P_n\!\left[\frac{A-\mu S_1(O)}{p_1}\right]+O_P(n^{-1}).
\end{equation}
At the true nuisances $(A-\mu S_1(O))/p_1$ is the influence function
$\phi_{\mathrm{pDP},1}(\,\cdot\,;P)$. If both nuisances
are $\mathcal{L}_2$-consistent then
$\|(A-\mu S_1(O))/p_1-\phi_{\mathrm{pDP},1}(\,\cdot\,;P)\|=o_P(1)$, and because both
functions have mean zero and the evaluation sample is independent of
$\mathcal{T}$, the two empirical averages differ by $o_P(n^{-1/2})$. 
Together with $C=o_P(n^{-1/2})$, this gives the expansion required by Proposition \ref{prop:linear}.
 Applying the argument to both groups yields asymptotic normality of the contrast and consistency of its empirical influence-function variance; Wald coverage requires the limiting contrast variance to be positive.
 Double-robust consistency alone supplies neither the product rate nor this score convergence when one nuisance remains misspecified.

\section{Probabilistic Equal Opportunity Derivation}
\label{app:equal}

Appendix \ref{app:demographic-parity} derives the efficient influence function of probabilistic demographic parity in full.
Here we derive probabilistic equal opportunity; thresholded opportunity is treated separately in Appendix \ref{app:threshold_stability}.
For the probabilistic metric, work with the $G=1$ component of the estimand, writing $\Psi_{\mathrm{pEO}}(P)=\Psi_1(P)-\Psi_0(P)$ with $\Psi_g(P)=\E[D(X)\mid Y=1,G=g]$ and recovering the $G=0$ component by symmetry; perturb $P$ along the point-mass contamination path
$$P_t = (1-t)P + t\,\delta_{\tilde{o}}, \qquad \tilde{o} = (\tilde{x},\tilde{g},\tilde{y});$$
express the component as a ratio $\Psi_1(P)=a/b$; and apply the quotient rule \eqref{eq:ratio}.
The conditioning event changes $a$, $b$, and the weight in the perturbation term of \eqref{eq:adot}.

The estimand is
\begin{equation}
    \Psi_{\mathrm{pEO}}(P) = \E[D(X) \mid Y=1,G=1] - \E[D(X) \mid Y=1,G=0],
\end{equation}
whose first component has $a = \E[D(X)\ind\{Y=1,G=1\}]$ and $b = \Prob(Y=1,G=1)$.
The only change from Section \ref{sec:parity} is that the conditioning event $\{G=1\}$ is replaced by $\{Y=1,G=1\}$.
Contamination therefore gives $\dot{b} = \ind\{\tilde{y}=1,\tilde{g}=1\} - b$ and
$$\dot{a} = \ind\{\tilde{y}=1,\tilde{g}=1\}D(\tilde{x}) - a + \E\left[\rho_1(X)\,\dot{D}(X)\right],$$
where the identity $\E[\ind\{Y=1,G=1\}h(X)] = \E[\rho_1(X)h(X)]$, with $\rho_1(x) = \Prob(Y=1,G=1\mid X=x)$, takes the place of the corresponding identity for $\pi$.
The pathwise derivative $\dot{D}$ is unchanged, so the perturbation term again picks out the contaminating point,
$$\E\left[\rho_1(X)\dot{D}(X)\right] = \int f_X(x)\,\rho_1(x)\,\frac{\delta_{\tilde{x}}(x)}{f_X(x)}\left(\ind\{\tilde y=1\} - D(x)\right)dx = \rho_1(\tilde{x})\left(\ind\{\tilde y=1\} - D(\tilde{x})\right),$$
where $f_X$ denotes the marginal density of $X$, and \eqref{eq:ratio} gives
\begin{equation}
    \phi_{\mathrm{pEO},1}(o;P) = \frac{1}{\Prob(Y=1,G=1)}\left[\rho_1(x)\,r(o) + \ind\{y=1,\,g=1\}\big(D(x) - \Psi_1(P)\big)\right],
\end{equation}
with $r$ the residual \eqref{eq:residual}.
Setting the sample mean of $\phi_{\mathrm{pEO},1}$ to zero and solving for $\Psi_1(P)$ yields
\begin{equation}
    \bar{\Psi}_{\mathrm{pEO},1}(\hat{P}) = \frac{1}{n\,\hat{\Prob}(Y=1,G=1)}
    \sum_{i=1}^n \Big[\hat{\rho}_1(x_i)\,\hat{r}_i + \ind\{y_i=1,\,g_i=1\}\,\hat{D}(x_i)\Big],
\end{equation}
with $\hat{r}_i = \ind\{y_i=1\} - \hat{D}(x_i)$ the fitted residual.

\section{Conditional Mutual Information EIF Derivation}
\label{app:cmi_eif}

For $O=(X,G,Y)$ with binary $G$ and $Y$, write
\[
  \Psi_{\mathrm{CMI}}(P)=P[\ell(O;P)],\qquad
  \ell(O;P)=\log\frac{p(G,Y\mid X)}{p(G\mid X)p(Y\mid X)},
\]
with both margins obtained from the same joint conditional law.
Assume these probabilities are positive, $\ell$ is square-integrable, and differentiation can be interchanged with expectation.
Consider a smooth, dominated submodel $P_t$ through $P$, with score $s(O)=\left.\frac{d}{dt}\log\frac{dP_t}{dP}(O)\right|_{t=0}$ satisfying $P[s(O)]=0$.
A dot denotes differentiation at $t=0$.
The product rule gives
\begin{equation}
  \left.\frac{d}{dt}\Psi_{\mathrm{CMI}}(P_t)\right|_{t=0}
  =P[\ell(O;P)s(O)]+P[\dot\ell(O)],
  \label{eq:cmi_path_derivative}
\end{equation}
where
\[
  \dot\ell(O)
  =\frac{\dot p(G,Y\mid X)}{p(G,Y\mid X)}
   -\frac{\dot p(G\mid X)}{p(G\mid X)}
   -\frac{\dot p(Y\mid X)}{p(Y\mid X)}.
\]
Each term has conditional mean zero: for any of the three discrete variables $A=(G,Y),G,Y$,
\[
  \E_P\!\left[\frac{\dot p(A\mid X)}{p(A\mid X)}\,\middle|\,X=x\right]
  =\sum_a\dot p(a\mid x)
  =\left.\frac{d}{dt}\sum_a p_t(a\mid x)\right|_{t=0}=0.
\]
Thus $P[\dot\ell(O)]=0$, and \eqref{eq:cmi_path_derivative} becomes
\[
  \left.\frac{d}{dt}\Psi_{\mathrm{CMI}}(P_t)\right|_{t=0}
  =P\!\left[\{\ell(O;P)-\Psi_{\mathrm{CMI}}(P)\}s(O)\right].
\]
The resulting mean-zero, square-integrable gradient is the EIF in the unrestricted nonparametric model:
\begin{equation}
  \phi(o;P)=\log\frac{p(g,y\mid x)}{p(g\mid x)p(y\mid x)}-\Psi_{\mathrm{CMI}}(P).
  \label{eq:cmi_eif_appendix}
\end{equation}
The one-step correction therefore simplifies to
\[
  \Psi_{\mathrm{CMI}}(\hat P)+P_n\phi(O;\hat P)
  =\Psi_{\mathrm{CMI}}(\hat P)+P_n\{\ell(O;\hat P)-\Psi_{\mathrm{CMI}}(\hat P)\}
  =P_n\ell(O;\hat P),
\]
as in \eqref{eq:cmi_est}; its empirical centering does not remove the higher-order remainder.

For TMLE, let $q_c(x)=p(g,y\mid x)$, where $c=(g,y)$ indexes the four joint states, and define
\[
  \begin{aligned}
    L_{q,c}(x)&=\log\frac{q_c(x)}{q_G(g\mid x)q_Y(y\mid x)},\\
    h_q(x)&=\sum_cq_c(x)L_{q,c}(x),\qquad
    H_{q,c}(x)=L_{q,c}(x)-h_q(x).
  \end{aligned}
\]
Since $\sum_cq_c(x)H_{q,c}(x)=0$, the exponential tilt
\begin{equation}
  q_{\epsilon,c}(x)=\frac{q_c(x)\exp\{\epsilon H_{q,c}(x)\}}
  {\sum_dq_d(x)\exp\{\epsilon H_{q,d}(x)\}}
  \label{eq:cmi_fluctuation}
\end{equation}
has conditional likelihood score $\sum_iH_{q,C_i}(x_i)$ at zero.
At each iteration, we choose a step along this family that increases the conditional log likelihood of the observed joint outcomes, recompute $H_q$, and repeat until the empirical score is approximately zero.
The final substitution estimate $P_nh_{q^*}(X)$ is nonnegative and satisfies
\[
  P_n\phi(O;P^*)=P_n\{L_{q^*,C}(X)-h_{q^*}(X)\}\approx0.
\]
It therefore approximately equals the observed log-ratio average at the targeted law, whereas the one-step estimator retains the initial law.

Targeting uses evaluation labels, so independence of the initial nuisance fit does not by itself control the updated empirical-process term.
Along with influence-function consistency and the integrability conditions in Proposition \ref{prop:linear}, sufficient conditions for asymptotic linearity are
\[
  \begin{aligned}
    P_n\phi(O;P^*)&=o_P(n^{-1/2}),\\
    (P_n-P)\{\phi(O;P^*)-\phi(O;P)\}&=o_P(n^{-1/2}),\\
    R(P^*,P)&=o_P(n^{-1/2}).
  \end{aligned}
\]
Wald inference additionally requires positive limiting EIF variance.
At conditional independence, $\ell(O;P)=\Psi_{\mathrm{CMI}}(P)=0$, so $\phi(O;P)=0$ almost surely; the first-order expansion is degenerate and does not establish Wald calibration at this null.

\section{Thresholded Metrics: Derivation and Stability}
\label{app:threshold_stability}

Thresholded inference requires an additional argument to show that we can control observations near the decision boundary.
The condition $\Prob(D(X)=c)=0$ alone does not provide this control: a small change in $D$ can move observations across $c$ even when none lie exactly there.
We instead bound the probability that the fitted and true decisions disagree.

\subsection{Estimands and estimators}

Fix a threshold $c\in(0,1)$ and let $D(x)=\Prob(Y=1\mid X=x)$.
Write $D_c(x)=\ind\{D(x)\geq c\}$ and $\hat D_c(x)=\ind\{\hat D(x)\geq c\}$ for the true and fitted decisions.
Let $S_g(O)=\ind\{G=g\}$ for parity and $S_g(O)=\ind\{Y=1,G=g\}$ for equal opportunity, where $g\in\{0,1\}$.
In either case define
\[
  \begin{aligned}
    p_g&=\E[S_g(O)],\qquad
    \Psi_g(P)=\E[D_c(X)\mid S_g(O)=1], \\
    \Psi(P)&=\Psi_1(P)-\Psi_0(P).
  \end{aligned}
\]
Thus $\Psi(P)$ is the corresponding thresholded data-fairness estimand.
Fit $\hat D$ on $m$ training observations, independently of $n$ i.i.d. evaluation observations $O_1,\ldots,O_n\sim P$.
As in Section \ref{sec:setup}, write $P_n[f(O)]=n^{-1}\sum_{i=1}^n f(O_i)$ and set
\[
  \begin{aligned}
    \hat p_g&=P_n[S_g],\qquad
    \bar{\Psi}_g(\hat{P})=\frac{P_n[S_g\hat D_c]}{\hat p_g}, \\
    \bar{\Psi}(\hat{P})&=\bar{\Psi}_1(\hat{P})-\bar{\Psi}_0(\hat{P}).
  \end{aligned}
\]
Here $\hat p_g=\hat{\Prob}(S_g(O)=1)$ is the empirical stratum probability. Set a component estimate to zero if its evaluation stratum is empty; this event has vanishing probability under the conditions below.

\subsection{Stability assumptions and asymptotic linearity}

The margin condition below adapts the classification margin assumption of \citet{audibertFastLearningRates2007} to threshold $c$ and each evaluation stratum. Their Lemma 5.1 bounds decision disagreement away from exact threshold ties by a power of the supremum-norm regression error. Our condition additionally excludes mass at $D(X)=c$; we use the resulting disagreement bound to establish asymptotic linearity of the fairness contrast.

\begin{proposition}[threshold stability]
  \label{prop:threshold_stability}
  Suppose $p_g\geq p_*>0$ for both groups.
  For constants $C,\kappa,\epsilon_0>0$, assume
  \begin{equation}
    \Prob\bigl(|D(X)-c|\leq\epsilon\mid S_g(O)=1\bigr)
    \leq C\epsilon^\kappa,
    \qquad 0<\epsilon\leq\epsilon_0,\quad g\in\{0,1\}.
    \label{eq:margin}
  \end{equation}
  Let $m=m(n)\to\infty$ as $n\to\infty$, and suppose that for a positive deterministic sequence $r_m\to0$,
  \begin{equation}
    \|\hat D-D\|_\infty=O_P(r_m),\qquad
    r_m^\kappa=o(n^{-1/2}),
    \label{eq:threshold_rate}
  \end{equation}
  where $\|\cdot\|_\infty$ is the essential supremum with respect to the marginal law of $X$.
  Then
  \begin{equation}
    \begin{aligned}
      \bar{\Psi}(\hat{P})-\Psi(P)&=P_n[\phi(O;P)]+o_P(n^{-1/2}), \\
      \phi(O;P)&=\frac{S_1(O)}{p_1}\{D_c(X)-\Psi_1(P)\}
      -\frac{S_0(O)}{p_0}\{D_c(X)-\Psi_0(P)\}.
    \end{aligned}
    \label{eq:threshold_linear}
  \end{equation}
  If $\sigma^2=\Var_P[\phi(O;P)]>0$, the interval
  $\bar{\Psi}(\hat{P})\pm z_{1-\alpha/2}\hat\sigma/\sqrt n$ has limiting coverage $1-\alpha$ for $\Psi(P)$, where $0<\alpha<1$, $z_q$ is the standard normal $q$-quantile, and $\hat\sigma^2$ is the sample variance of
  \[
    \begin{aligned}
      \phi(O_i;\hat{P})&=\frac{S_1(O_i)}{\hat p_1}\{\hat D_c(X_i)-\bar{\Psi}_1(\hat{P})\} \\
      &\quad -\frac{S_0(O_i)}{\hat p_0}\{\hat D_c(X_i)-\bar{\Psi}_0(\hat{P})\}.
    \end{aligned}
  \]
\end{proposition}

\begin{proof}
  Let $\Delta_m=\|\hat D-D\|_\infty$.
  If $\hat D_c(X)\neq D_c(X)$, then $|D(X)-c|\leq\Delta_m$.
  Conditional on the training data, the disagreement probability $q_g$ in stratum $g$ therefore satisfies, on the event $\Delta_m\leq\epsilon_0$,
  \[
    q_g\leq C\Delta_m^\kappa=O_P(r_m^\kappa)=o_P(n^{-1/2}).
  \]
  The event has probability tending to one.
  The conditional expectation of $P_n[S_g|\hat D_c-D_c|]$ is $p_gq_g$ by independence of the evaluation sample, so conditional Markov's inequality gives
  $P_n[S_g|\hat D_c-D_c|]=o_P(n^{-1/2})$.
  Since $\hat p_g\to_P p_g>0$, the fitted-rule estimate differs negligibly from the oracle estimate $\tilde{\Psi}_g=P_n[S_gD_c]/\hat p_g$:
  \[
    |\bar{\Psi}_g(\hat{P})-\tilde{\Psi}_g|
    \leq\frac{P_n[S_g|\hat D_c-D_c|]}{\hat p_g}=o_P(n^{-1/2}).
  \]
  The oracle ratio satisfies
  \[
    \tilde{\Psi}_g-\Psi_g(P)
    =\frac{P_n[S_g(D_c-\Psi_g(P))]}{\hat p_g}
    =P_n\!\left[\frac{S_g}{p_g}(D_c-\Psi_g(P))\right]+O_P(n^{-1}),
  \]
  because its numerator and $\hat p_g-p_g$ are both $O_P(n^{-1/2})$.
  Subtracting the two group expansions proves \eqref{eq:threshold_linear}.
  The summands are bounded and mean-zero, so the central limit theorem applies.
  Finally, the vanishing empirical disagreement and consistency of $\hat p_g$ and $\bar{\Psi}_g(\hat{P})$ give
  $P_n[(\phi(O;\hat{P})-\phi(O;P))^2]=o_P(1)$, hence $\hat\sigma^2\to_P\sigma^2$.
  Slutsky's theorem gives the stated coverage.
\end{proof}

\subsection{Interpreting the assumptions}

The margin exponent and training rate must be considered together.
A bounded conditional density of $D(X)$ near $c$ gives \eqref{eq:margin} with $\kappa=1$, requiring $r_m=o(n^{-1/2})$; it does not by itself justify the intervals.
With $\kappa=2$, the sufficient rate is $r_m=o(n^{-1/4})$ in the \emph{supremum} norm, not merely $\mathcal L_2$.
A gap $|D(X)-c|\geq\delta>0$ in both strata is a simpler special case: uniform consistency makes $\hat D_c=D_c$ there with probability tending to one, so no relative rate between $m$ and $n$ is needed.
The proposition establishes pointwise asymptotic inference for the true thresholded rule under these stability conditions; it does not assert an efficient influence function or regularity in an unrestricted nonparametric model.

\subsection{Thresholded equal opportunity}

For thresholded equal opportunity the estimand is
\begin{equation}
    \Psi_{\mathrm{EO}}(P) = \E[D_c(X)\mid Y=1,G=1] - \E[D_c(X)\mid Y=1,G=0],
\end{equation}
Take $S_g(O)=\ind\{Y=1,G=g\}$ in Proposition \ref{prop:threshold_stability}.
Under its positivity, margin, and training-rate conditions, the group-specific influence representation is
\begin{equation}
    \phi_{\mathrm{EO},1}(o;P) = \frac{\ind\{y=1,\,g=1\}}{\Prob(Y=1,G=1)}\left(D_c(x) - \Psi_1(P)\right),
\end{equation}
where $\Psi_1(P)=\E[D_c(X)\mid Y=1,G=1]$.
The corresponding estimator is
\begin{equation}
    \bar{\Psi}_{\mathrm{EO},1}(\hat{P}) = \frac{1}{n\,\hat{\Prob}(Y=1,G=1)}
    \sum_{i=1}^n \ind\{y_i=1,\,g_i=1\}\,\hat{D}_c(x_i).
\end{equation}
Subtracting the group-0 component gives the opportunity contrast; the variance is the variance of the difference of the two representations on all evaluation observations, including zeros when $Y=0$.

\section{Nuisance Estimation and Specification Checks}
\label{app:nuisance-simulations}

We study how training size and learner capacity affect inference for probabilistic demographic parity.
We also compare true conditional probabilities with correctly specified and misspecified fitted models to distinguish bias correction from confidence-interval calibration.

\subsection{Simulation Design}
\label{app:nuisance-design}

Let $X=(U,V,W_1,W_2,W_3)$ have independent standard-normal components, with $U$ shared by group membership and the outcome, $V$ predicting only the outcome, and $W_j$ containing no signal.
Draw $G$ and $Y$ independently given $X$, with
\[
  \begin{gathered}
    h_\lambda(z)=\sqrt{1-\lambda^2}\,z+\lambda(z^2-1)/\sqrt{2},\\
    \pi(X)=\expit\{h_\lambda(U)\},\qquad
    D(X)=\expit\{h_\lambda(V)+\delta h_\lambda(U)\},
  \end{gathered}
\]
where $\lambda=0.5$ and $\delta=0.6$.
We use $R=1000$ independent replicates per configuration and 2000 evaluation observations.
The boosting experiment uses training sizes $100, 250, 500, 1000, 2500, 5000,$ and $10{,}000$.
Within each replicate, comparisons use the same simulated data, and each fitted outcome or group model is reused across comparisons involving that model.
Population parity is calculated by eight independently scrambled Sobol integrations, increasing resolution until the integration standard error is below one percent of the smallest bias Monte Carlo standard error.

Both gradient-boosted classifiers in Figure~\ref{fig:trainsize} use raw features and learning rate 0.1, with maxima of 100 depth-3 or 1000 depth-5 trees.
For each nuisance, a stratified 20\% of training observations is reserved for validation; training stops after 20 iterations without improvement of at least $10^{-4}$, without refitting on the full training sample.
For each nuisance, the lower- and higher-capacity classifiers use the same validation observations; evaluation data never enter fitting or stopping decisions.

\subsection{Specification and Interval Calibration}

The product-bias identity in \eqref{eq:product} allows one nuisance to compensate for error in the other.
To examine how this property translates into inference, we hold the population fixed and vary the outcome and group nuisances separately, comparing point-estimation bias with interval coverage.

We use three procedures for each nuisance.
Oracle supplies the true conditional probabilities.
Quadratic fits logistic regression with an intercept and raw and squared features, so its model class contains the true nuisance function.
Linear omits the squared features and is therefore misspecified.
Both fitted procedures use negligible regularization, with inverse regularization strength $C=10^8$.
We apply each procedure to both nuisances and pair Oracle or Quadratic with Linear in both directions.
The oracle comparisons assess double robustness without fitting error in the correct nuisance; the quadratic fits introduce training uncertainty within a correct model class.

These specification checks use training sizes $500, 2500,$ and $10{,}000$, with the same evaluation size and number of replicates as the boosting experiment.
Table~\ref{tab:nuisance-specification} reports signed bias and coverage of nominal 95\% Wald intervals for population probabilistic demographic parity at the smallest and largest of these training sizes.
The oracle comparisons have small signed bias and near-nominal coverage even when the other nuisance is misspecified.
Correctly specified fits also give small bias, but their intervals under-cover at small training sizes.
Coverage improves with more training data in these cases; when both fits are linear, bias and undercoverage persist.

For $R=1000$ independent replicates per configuration, coverage and signed-bias Monte Carlo standard errors are $\sqrt{\hat c(1-\hat c)/R}$ and $s_{\hat\Psi}/\sqrt R$, respectively, where $\hat c$ is the proportion of intervals containing the true population value and $s_{\hat\Psi}$ is the empirical standard deviation across replicate estimates.

\begin{table}[t]
\centering\small
\caption{TL coverage and signed bias for $R=1000$ replicates per configuration, at the smallest and largest training sizes in the nuisance-specification experiment. Intervals have nominal 95\% coverage. Parentheses contain Monte Carlo standard errors; bias is in percentage points (pp).}
\label{tab:nuisance-specification}
\begin{tabular}{@{}llrrrr@{}}
\toprule
 & & \multicolumn{2}{c}{$n_{\rm train}=500$} & \multicolumn{2}{c}{$n_{\rm train}=10{,}000$}\\
\cmidrule(lr){3-4}\cmidrule(l){5-6}
Outcome $\hat D$ & Group $\hat\pi$ & Coverage & Bias (pp) & Coverage & Bias (pp)\\
\midrule
Oracle & Oracle & 0.952 (0.007) & 0.01 (0.04) & 0.951 (0.007) & 0.03 (0.04)\\
Oracle & Linear & 0.949 (0.007) & 0.01 (0.04) & 0.951 (0.007) & 0.03 (0.04)\\
Linear & Oracle & 0.953 (0.007) & 0.03 (0.04) & 0.955 (0.007) & 0.04 (0.04)\\
\midrule
Quadratic & Quadratic & 0.917 (0.009) & -0.01 (0.04) & 0.950 (0.007) & 0.03 (0.04)\\
Quadratic & Linear & 0.903 (0.009) & 0.02 (0.05) & 0.950 (0.007) & 0.03 (0.04)\\
Linear & Quadratic & 0.873 (0.011) & 0.03 (0.05) & 0.947 (0.007) & 0.03 (0.04)\\
Linear & Linear & 0.679 (0.015) & -1.58 (0.04) & 0.720 (0.014) & -1.56 (0.04)\\
\bottomrule
\end{tabular}
\end{table}

\FloatBarrier

\section{CMI Simulation}
\label{app:cmi}

\begin{figure}
  \centering
  \includegraphics[width=0.45\linewidth]{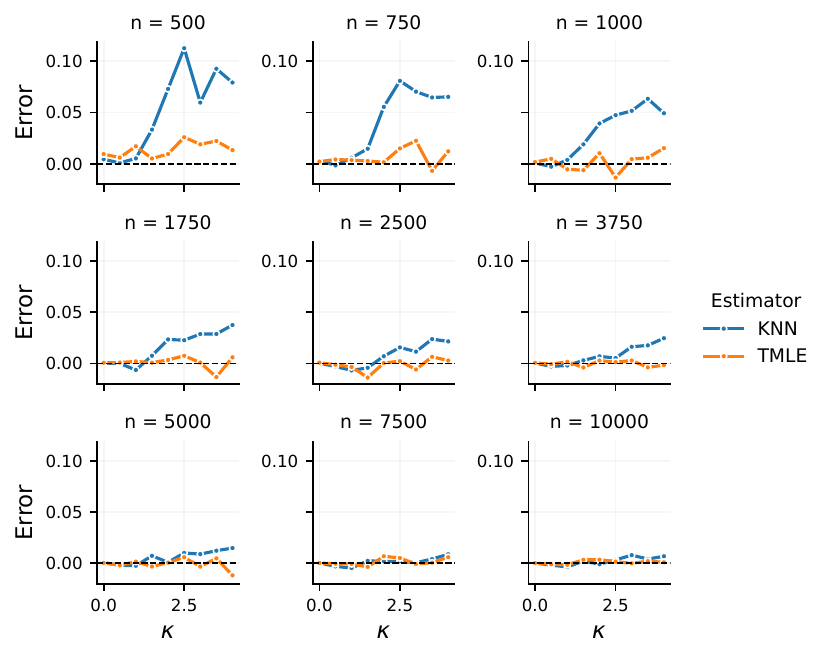}
  \includegraphics[width=0.5\linewidth]{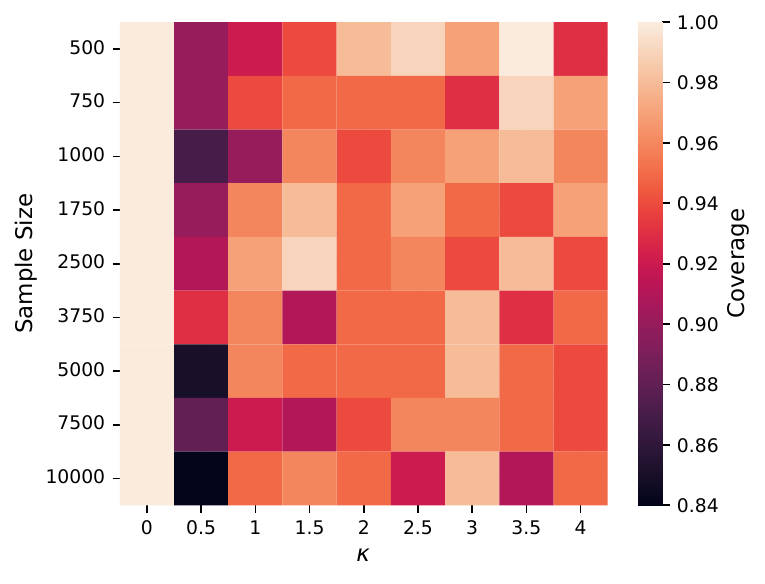}
  \caption{CMI estimation error for TMLE and KNN (left, 10 replicates per estimator and configuration) and nominal two-sided TMLE interval coverage (right, 100 replicates per configuration). Reference lines mark zero error.}
  \label{fig:cmi}
\end{figure}

We compare TMLE with the mixed discrete--continuous $k$-nearest-neighbor (KNN) estimator of CMI \cite{mesnerConditionalMutualInformation2019}, using KNN as a point-estimation baseline and simulation truth to assess interval coverage.
For all CMI analyses, we estimate the joint group--outcome probabilities using gradient-boosted decision trees of depth 3, learning rate 0.1, and at most 100 boosting rounds, with one tree per joint-response class in each round.
A stratified 20\% of the training observations is reserved for validation; training stops after 20 rounds without improvement of at least $10^{-4}$ in validation log loss, with no probability-calibration step or full-training refit.
For the one-step and TMLE simulations, the displayed sample size $n$ is split equally between training and evaluation.
KNN uses all $n$ observations with $k=7$ neighbors and is evaluated on independently drawn datasets from the same distribution.
Draw $X\sim N(0,I_3)$ and independent $S,U,V\sim\Uniform(0,1)$, and set
\[
  Y=\ind\!\left\{\frac{\kappa S+U+\expit(\beta^TX)}{\kappa+2}>\frac12\right\},\qquad
  G=\ind\!\left\{\frac{\kappa S+V+\expit(\beta^TX)}{\kappa+2}>\frac12\right\},
\]
with $\beta=(1,1,1)^T$.
The shared $S$ induces conditional dependence for $\kappa>0$; at $\kappa=0$, $Y\independent G\mid X$ and the true CMI is exactly zero.
We use coupling weights $\kappa=0,0.5,\ldots,4$ and sample sizes $n=500,750,1000,1750,2500,3750,5000,7500,$ and $10{,}000$.
For each configuration, the KNN comparison uses 10 replicates per estimator; a separate set of 100 replicates is used to assess coverage.
Table \ref{tab:cmi_truth} gives Monte Carlo conditional CMI truth for the remaining coupling weights, averaging over $10^6$ draws of $X$ and 2048 draws of $S$ per feature vector, with $U$ and $V$ integrated out analytically.
Figure \ref{fig:cmi} shows estimation error and coverage as coupling and sample size vary.
Coverage remains sensitive to coupling and sample size even with validation-based stopping.

\begin{table}
  \centering
  \begin{tabular}{c|c}
$\kappa$ & $I(Y;G\mid X)$ \\ \hline
0 & 0.0000 \\
0.5 & 0.0054 \\
1 & 0.0420 \\
1.5 & 0.1189 \\
2 & 0.2080 \\
2.5 & 0.2770 \\
3 & 0.3274 \\
3.5 & 0.3660 \\
4 & 0.3964 \\
\end{tabular}

  \caption{Monte Carlo conditional CMI truth $I(Y;G\mid X)$ by coupling weight; the value at $\kappa=0$ is exactly zero.}
  \label{tab:cmi_truth}
\end{table}

The conditioning-set experiment in Section \ref{sec:conditioning} uses 30 independent standard-normal features: five shared confounders, eight outcome-only predictors, and 17 noise features.
Coefficient magnitudes are drawn independently from $1.5\,\Uniform(0.5,1.5)$, with equally likely positive or negative signs; each confounder has the same sign in the group and outcome equations.
The coefficients are drawn once with seed 0 and held fixed across replicates and conditioning sets.
We include the first $k=0,\ldots,5$ confounders in their original column order, always retaining the outcome-only and noise features.
Each configuration uses $R=200$ independent replicates at total sample sizes $n=2500$ and $5000$, with new datasets drawn for each conditioning set.
Both simulation designs use master seed 123.

\begin{figure}
  \centering
  \includegraphics[width=\linewidth]{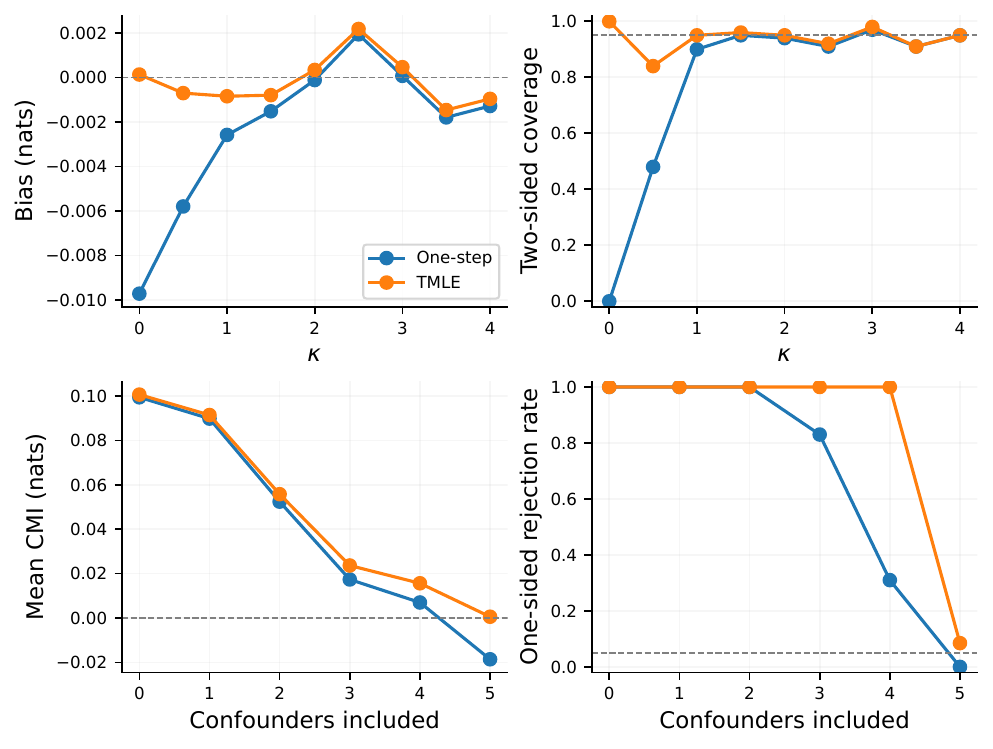}
  \caption{Paired one-step and TMLE results at the largest sample size in each experiment. Top: bias and two-sided coverage as coupling varies ($n=10{,}000$, $R=100$). Bottom: mean CMI and one-sided rejection rates as confounders are added ($n=5000$, $R=200$). Dashed lines mark zero, nominal coverage, or the nominal rejection cutoff.}
  \label{fig:cmi_paired}
\end{figure}

To isolate targeting, we also compute one-step estimates from the same initial joint-classifier fits, splits, and evaluation observations as TMLE within each replicate.
Figure \ref{fig:cmi_paired} compares the methods in both simulation designs.
Table \ref{tab:cmi_null_diagnostics} summarizes their null calibration at the largest sample size in each design, with Monte Carlo standard errors.
These empirical diagnostics assess the nominal Wald rules; targeting does not change the zero-EIF degeneracy derived in Appendix \ref{app:cmi_eif}.

\begin{table}
  \centering
  \small
\begin{tabular}{@{}lrlrrr@{}}
\toprule
Experiment & $n$ & Method & Bias & Coverage & Rejection \\
\midrule
Conditioning & 5000 & One-step & -0.0186 (0.00039) & 0.03 (0.01) & 0.00 (0.00) \\
Conditioning & 5000 & TMLE & 0.000568 (4.4e-05) & 0.97 (0.01) & 0.09 (0.02) \\
Coupling & 10000 & One-step & -0.00971 (0.0002) & 0.00 (0.00) & 0.00 (0.00) \\
Coupling & 10000 & TMLE & 0.000128 (1.5e-05) & 1.00 (0.00) & 0.00 (0.00) \\
\bottomrule
\end{tabular}

  \caption{Empirical bias, two-sided coverage, and one-sided rejection at conditional independence, using the largest sample size in each design. Parentheses contain Monte Carlo standard errors.}
  \label{tab:cmi_null_diagnostics}
\end{table}

\section{Data Analysis Details}
\label{app:data-details}

The Adult-Income input is \href{https://github.com/socialfoundations/folktables/blob/ebace19d8065efc43589f0067bf183398eb156b9/adult_reconstruction.csv}{\texttt{adult\_reconstruction.csv}} from the folktables repository (revision \texttt{ebace19}); the Law School input is \href{https://github.com/tailequy/fairness_dataset/blob/4d225df30598951985c31d81e1544dba504286a8/experiments/data/law_school_clean.csv}{\texttt{law\_school\_clean.csv}} from the dataset collection accompanying Le Quy et al.'s survey (revision \texttt{4d225df}).
We retain all rows in these files and use a shuffled, unstratified 60/40 training/evaluation split with seed 123.
Within each dataset, all metrics use the same split and feature set (Table~\ref{tab:data-protocol}).

\begin{table}
  \centering
  \caption{Dataset sizes and features used in Section~\ref{sec:data}. The outcome $Y$ and group $G$ are excluded from $X$ in both datasets.}
  \label{tab:data-protocol}
  \begin{tabular}{@{}p{0.14\linewidth}p{0.22\linewidth}p{0.58\linewidth}@{}}
    \toprule
    Dataset & Observations & Features in $X$ \\
    \midrule
    Adult-Income & 49,531 total\newline 29,718 training\newline 19,813 evaluation & hours-per-week, age, capital-gain, capital-loss, workclass, education, marital-status, relationship, race, native-country, occupation \\
    \addlinespace
    Law School & 20,798 total\newline 12,478 training\newline 8320 evaluation & \texttt{decile1b}, \texttt{decile3}, \texttt{lsat}, \texttt{ugpa}, \texttt{zfygpa}, \texttt{zgpa}, \texttt{fulltime}, \texttt{fam\_inc}, \texttt{male}, \texttt{tier} \\
    \bottomrule
  \end{tabular}
\end{table}

For Adult-Income, we omit \texttt{education-num} and encode each categorical feature by consecutive integers in sorted category order, retaining hours worked, age, capital gain, and capital loss as numeric variables.
The category vocabulary is constructed from covariates in both the training and evaluation splits, without using evaluation outcomes or group labels.
For Law School, the supplied numeric features are retained, and the supplied race categories are encoded as White ($G=1$) and Non-White ($G=0$).
Neither input contains missing values requiring imputation; categorical values are retained as supplied.

Each SuperLearner ensemble contains six base learners: logistic regression, histogram gradient boosting, $k$-nearest neighbors, a support-vector classifier, a random forest, and a multilayer perceptron.
We use scikit-learn 1.9.0 defaults except that logistic regression allows 2000 iterations, the multilayer perceptron allows 1000 iterations with early stopping, and stochastic learners use seed 123.
The logistic regression, nearest-neighbor, support-vector, and neural-network learners use standardized inputs, with scaling fitted within each training fold.
The support-vector classifier uses sigmoid probability calibration based on three-fold cross-validation.
Ten shuffled, stratified folds of the outer training sample generate out-of-fold class probabilities from each base learner.
A logistic regression with at most 2000 iterations combines these probabilities; the base learners are then refitted on the outer training sample for evaluation predictions.
Separate ensembles estimate the outcome, group, and four-class joint group--outcome distributions.
CMI uses the boosted joint classifier and validation procedure specified in Appendix~\ref{app:cmi}, followed by the targeting update in Section~\ref{sec:cmi}.


\end{document}